%% file: main.tex
\documentclass[letterpaper, 10 pt, conference]{ieeeconf}  

\IEEEoverridecommandlockouts                              

\usepackage{ifthen}

\newcommand\blindmode{false} 

\usepackage{graphics} 
\usepackage{epsfig} 
\usepackage{times} 
\usepackage[cmex10]{amsmath}
\usepackage{amssymb}  

\makeatletter
\let\NAT@parse\undefined
\makeatother

\input{preamble.tex}
\input{macros.tex}

\newcommand{\red}[1]{} 

\title{\Large \bf
Quadrotor Navigation using Reinforcement Learning with Privileged Information
}
\ifthenelse{\equal{\blindmode}{true}}{
\author{Author names omitted for anonymous review}
}{
\author{Jonathan Lee, Abhishek Rathod, Kshitij Goel, John Stecklein, and Wennie Tabib%
\thanks{The authors are with the Robotics Institute, Carnegie Mellon University,
Pittsburgh, PA 15213 USA.
\texttt{jlee6@alumni.cmu.edu},
\{\texttt{arathod2, kgoel1,jsteckle,wtabib}\}\texttt{@andrew.cmu.edu}.}
}
}

\begin{document}

\maketitle
\thispagestyle{empty}
\pagestyle{empty}

\begin{abstract}
This paper presents a reinforcement learning-based quadrotor navigation method
that leverages efficient differentiable simulation, novel loss functions, and
privileged information to navigate around large obstacles. Prior learning-based
methods perform well in scenes that exhibit narrow obstacles, but struggle when
the goal location is blocked by large walls or terrain. In contrast, the
proposed method utilizes time-of-arrival (ToA) maps as privileged information
and a yaw alignment loss to guide the robot around large obstacles. The policy
is evaluated in photo-realistic simulation environments containing large
obstacles, sharp corners, and dead-ends. Our approach achieves an 86\% success
rate and outperforms baseline strategies by 34\%. We deploy the policy onboard a
custom quadrotor in outdoor cluttered environments both during the day and
night. The policy is validated across 20 flights, covering \SI{589}{\meter}
without collisions at speeds up to \SI{4}{\meter\per\second}.
\end{abstract}

\section{INTRODUCTION}
\input{content/introduction}

\section{RELATED WORK}
\input{content/related_work}

\section{METHODOLOGY\label{sec:part2_methodology}}
\input{content/methodology}

\section{SIMULATION EXPERIMENTS}
\input{content/simulation_experiments}

\section{HARDWARE EXPERIMENTS\label{sec:part2_hardware}}
\input{content/hardware_experiments}

\section{CONCLUSION}
\input{content/conclusion}

\input{content/acknowledgements}

{
 \balance
 \small{
 \bibliographystyle{IEEEtranN}
 \bibliography{refs}
 }
}

\end{document}

%% file: preamble.tex
\usepackage{microtype}
\usepackage{graphicx}
\usepackage{epigraph}
\usepackage{amsfonts}
\usepackage[numbers,sort]{natbib}
\usepackage{setspace}
\usepackage{layout}
\usepackage[subfigure]{tocloft}  

\usepackage{pgfplots}
\usepackage{pgfplotstable}
\usepackage{tikz}
\pgfplotsset{compat=1.9}
\usetikzlibrary{positioning}
\usetikzlibrary{backgrounds}
\usetikzlibrary{calc}
\usepgfplotslibrary{colorbrewer}  
\usepgfplotslibrary{statistics}

\usetikzlibrary{external,positioning}
\usepgfplotslibrary{external}
\usepackage[format=hang]{subfig}  
\usepackage{xcolor}  
\usepackage{algorithm}
\usepackage{algpseudocode} 
\usepackage{ctable}
\usepackage{multirow}

\usepackage{xspace} 
\usepackage[per-mode=symbol]{siunitx} 

\usepackage{float}
\newfloat{algorithm}{tbp}{lop}

\usepackage{hyperref}

\usepackage{adjustbox}  
\usepackage{booktabs}   
\usepackage{multirow}   
\usepackage{amssymb}    
\usepackage{pifont}     
\usepackage{balance}    
\usepackage{soul}       

\usepackage[capitalize]{cleveref}

\crefname{section}{Sect.}{Sects.}
\Crefname{section}{Section}{Sections}
\crefname{subsection}{Sect.}{Sects.}
\Crefname{subsection}{Section}{Sections}
\crefname{subsubsection}{Sect.}{Sects.}
\Crefname{subsubsection}{Section}{Sections}
\crefname{figure}{Fig.}{Figs.}
\Crefname{figure}{Figure}{Figures}

\usepackage[font=small]{caption} 

%% file: macros.tex
\newcommand{\chapternote}[1]{{%
  \let\thempfn\relax
  \footnotetext[0]{\emph{#1}}
}}

\newcommand{\cmark}{\textcolor{green!60!black}{\ding{51}}} 
\newcommand{\xmark}{\textcolor{red}{\ding{55}}} 

%% file: content/introduction.tex
Traditional navigation approaches decompose perception, planning, state
estimation, and control into separate tasks. However, end-to-end learning-based
methods, which use a neural network to convert raw sensor observations into
actions, are increasingly being deployed for high-speed autonomous flight
\cite{loquercio2021learning,zhang2025learning,lu2024you}. The advantages of
end-to-end approaches are reduced planning latency and lower computational
overhead, which enables deployment on lightweight and low-cost platforms.
However, these approaches either require large amounts of expert-labeled data or
struggle to find paths in challenging environments that exhibit large obstacles,
sharp corners, and dead ends.

To bridge these gaps in the state of the art, we extend the method
of~\citet{zhang2025learning} to enable navigation around large obstacles.
\red{while retaining the efficient differentiable simulation-based training pipeline.}%
Like~\citet{zhang2025learning}, our policy is reactive at deployment,
relying only on depth observations and state estimates. However, by training
with ToA maps as privileged information, our policy learns globally-aware
navigation behavior without requiring a ToA map at test time.
We provide the following contributions:
\begin{enumerate}
\item an objective function for predicting heading (i.e., yaw) that improves
navigation performance relative to state-of-the-art approaches in environments
that require changes in orientation (e.g., twisting passageways and sharp
corners);
\item a method to leverage a time-of-arrival map as privileged information
during training, enabling shortest-path navigation without an explicit map at
test time;
\red{\item an approach to reduce attitude control response latency by computing
angular rate setpoints from predicted actions;
\item an approach to mitigate the effects of inaccurate motor and inertial
parameter estimation by leveraging domain randomization; and}
\item approaches to bridge the sim-to-real gap via body rate attitude
control and domain randomization; and
\item extensive evaluation of the system in photo-realistic simulation and
hardware experiments\ifthenelse{\equal{\blindmode}{true}}{}{\footnote{A video of
the experiments may be found at \url{https://youtu.be/RbHJ69o-zUc}.}} as well as an
open-source software release\ifthenelse{\equal{\blindmode}{true}}{\footnote{A link will
be added here upon acceptance of this
paper.}}{\footnote{\url{https://github.com/rislab/depthnav}}}.
\end{enumerate}

\begin{figure}
      \centering
      \includegraphics[width=0.99\linewidth,trim=20 0 100 0,clip]{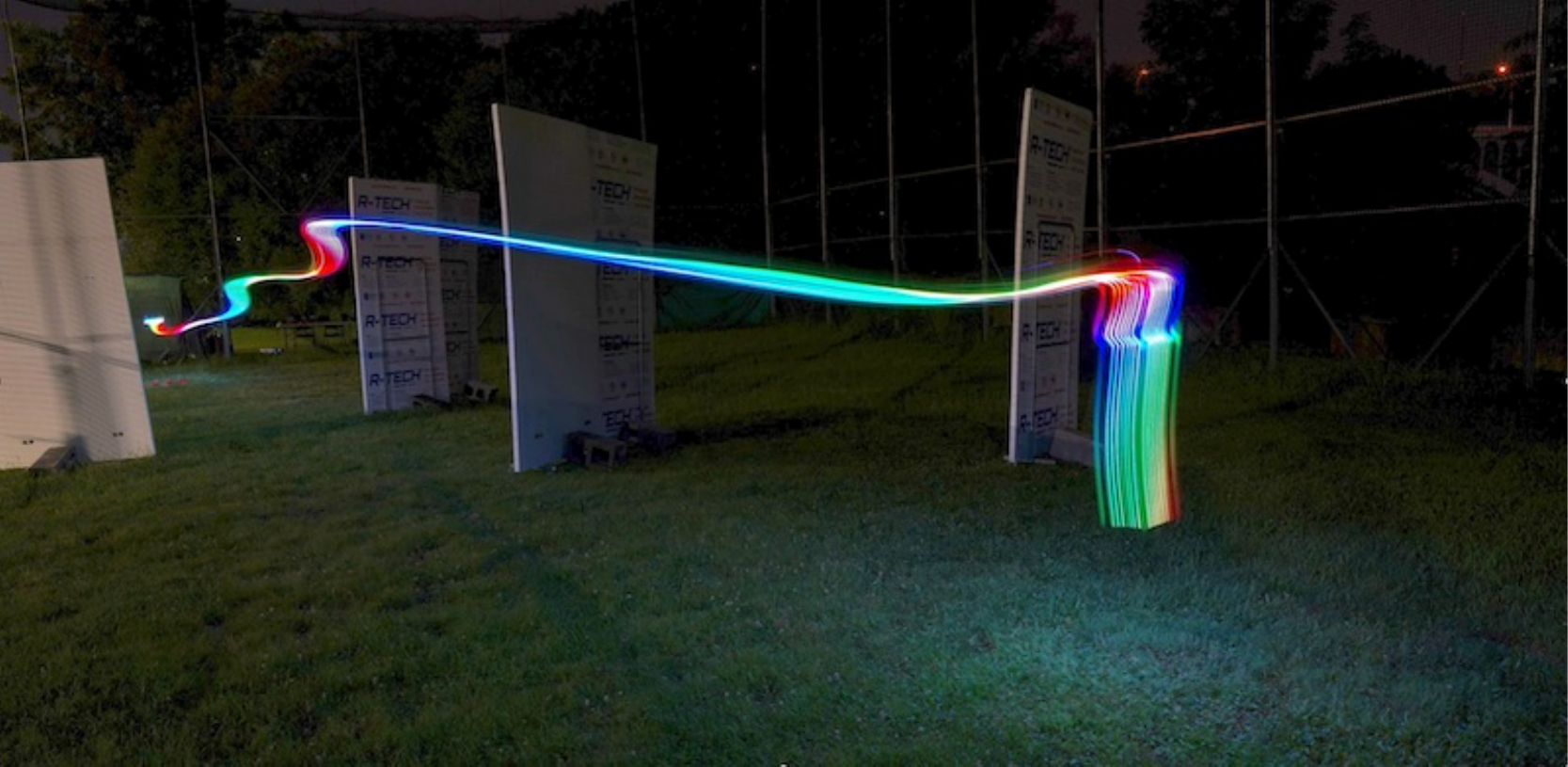}
      \caption{\label{fig:intro}
            This paper develops and deploys an end-to-end policy to navigate in
            challenging environments. The approach outperforms the state of the
            art by 34\%. An example trajectory is captured using long-exposure
            photography.}
      \vspace{-0.25cm}
\end{figure}

%% file: content/related_work.tex
\citet{loquercio2021learning} demonstrate high-speed, learning-based navigation
by training a student policy in simulation using a privileged, sample-based
expert planner. The student learns to predict control points of a polynomial
trajectory based on top-ranked expert samples. This approach requires a large
amount of expert-labeled data and increases the training overhead when the
expert planner is computationally intensive.~\citet{kim2025rapid} present an
Inverse Reinforcement Learning (IRL) method that similarly relies on expert
demonstrations, but instead of directly imitating the expert, a reward function
is inferred that explains the observations. Empirically, this approach enables
out-of-distribution generalization. However, the learnt policy lacks temporal
awareness during inference, making it susceptible to collisions when navigating
around large obstacles.

An alternative to relying on expert data is to train a policy by backpropagating
reward gradients through a differentiable simulator~\citep{newbury2024review}.
~\citet{zhang2025learning} leverage this idea by training a recurrent neural
network control policy with a lightweight point-mass dynamics model and
rendering engine. This method enables reliable operation at high speeds in
several environments.  However, the learnt policy does not generalize to
scenarios where significant yaw motion is required.

\red{In classical global path planning, occupancy grids or Euclidean signed distance
fields (ESDF) are used to avoid getting stuck in local extrema. In contrast, few}
Recent
learning-based navigation methods exploit geometric maps as privileged information
available during training but absent at inference~\citep{meijer2025pushing,
lu2024you}.  YOPO~\citep{lu2024you} trains a neural policy to generate motion
primitive offsets and scores using a Euclidean signed distance field
(ESDF) cost map as supervision. However,
this approach provides only limited guidance for escaping large convex obstacle
regions.

\red{Our approach improves upon the work of~\citet{zhang2025learning} by
incorporating a yaw prediction objective to navigate in
complex environments that require sudden changes in orientation (e.g, sharp
corners, walls, dead ends, etc.). In contrast to~\citep{lu2024you}, we enable
robustness to local extrema via privileged information about ToA
maps to goal positions without using a high-cost training simulator. In
addition, both~\citep{zhang2025learning,lu2024you} assume perfect system
knowledge for zero-shot sim-to-real transfer.  Conversely, our approach ensures
robustness to incorrect motor or inertial parameter identification via domain
randomization during training.}
Our approach addresses two limitations of prior work.
First, \citet{zhang2025learning} maintains a fixed heading toward the target,
limiting navigation around large obstacles. Our yaw alignment loss directly supervises heading
prediction, enabling the robot to yaw while changing direction.
Second, as ESDFs encode obstacle distance rather than path direction,
\citet{lu2024you} provides local avoidance but not globally optimal routing.
Our ToA maps address this by guiding the robot along shortest paths.
Together, the yaw alignment loss enables reorientation while the ToA gradient
teaches the robot which direction to head to make progress towards the goal.

%% file: content/methodology.tex
The proposed navigation policy takes a depth image, target
information, and the quadrotor state as inputs and uses a neural
network to predict a thrust and yaw angle.

\subsection{Differentiable Dynamics\label{ssec:diff_dynamics}}
\begin{figure}
  \centering
  \includegraphics[width=\linewidth,trim=32 20 32 18,clip]{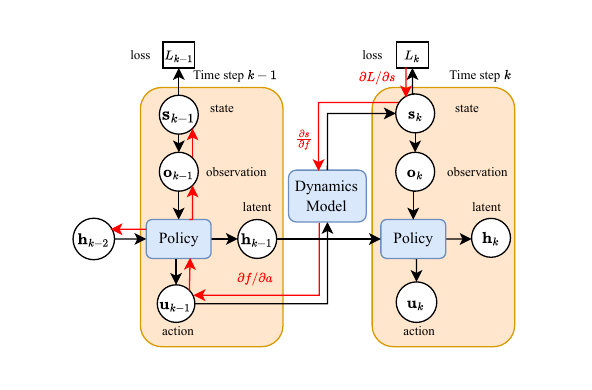}
  \caption{\label{fig:differentiable_sim} Differentiable dynamics enables direct
  policy updates by performing gradient descent on the loss function.
  }
  \vspace{-0.25cm}
\end{figure}

Following the approach of \citet{zhang2025learning}, the quadrotor navigation
problem is formulated as a Markov decision process with a discrete-time
dynamical system. As shown in
\cref{fig:differentiable_sim}, observations, $\mathbf{o}_k$, are generated at
each time step from sensor measurements based on the current state,
$\mathbf{s}_k$. The policy takes the observation $\mathbf{o}_k$ and a
hidden state $\mathbf{h}_{k-1}$ as inputs and outputs a new hidden state
$\mathbf{h}_k$ and action $\mathbf{u}_k$. The action $\mathbf{u}_k \in \mathbb{R}^4$
contains the mass-normalized thrust vector $\mathbf{t}_k \in \mathbb{R}^3$ and a
predicted yaw angle $\psi_k$, the latter is absent
in~\citep{zhang2025learning}. The action $\mathbf{u}_k$ is the control input to
the system dynamics $\mathbf{s}_{k+1} = f(\mathbf{s}_k, \mathbf{u}_k)$, which
determines the next state, $\mathbf{s}_{k+1}$.  At each time step $k$, the agent
also receives a loss value $L_k$ dependent on the current state and action.

Notably, the system dynamics are differentiable with respect to $\mathbf{s}_k$
and $\mathbf{u}_k$, enabling Analytical Policy Gradient (APG) methods to train
the model by backpropagating the loss through the
dynamics~\citep{freeman2021brax, newbury2024review}. This allows for sample
efficient training, as even a single sample provides a usable gradient for
policy optimization.
\begin{figure*}
  \centering
  \includegraphics[width=\textwidth,trim=20 9 35 9,clip]{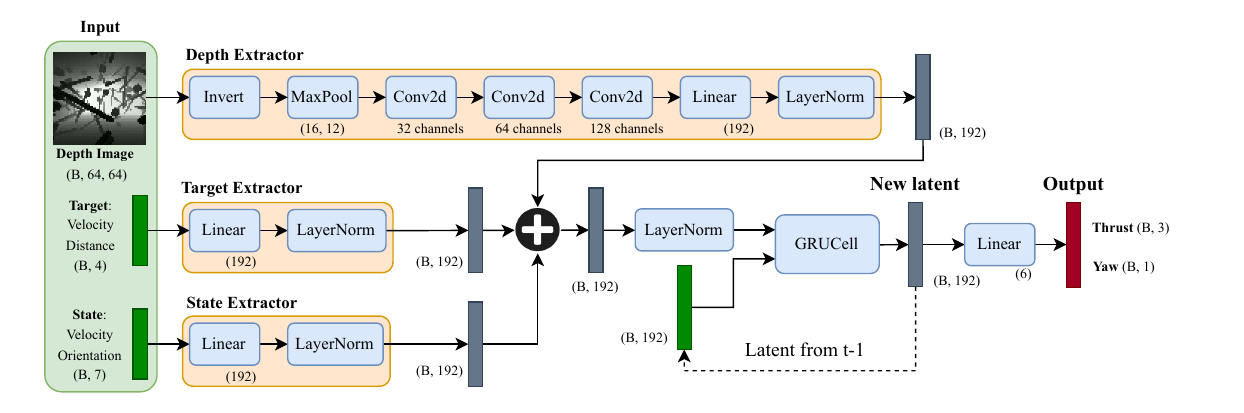}
  \caption{\label{fig:network_architecture}The end-to-end planning and control
    architecture is trained as a single neural network. Feature extractors
    process each input before they are flattened and summed together. A GRUCell
    helps to maintain consistent action predictions over time.}
  \vspace{-0.25cm}
\end{figure*}

While the full quadrotor dynamics model is differentiable, it has been shown
that using point-mass dynamics results in reduced training time without
sacrificing performance \citep{heegLearningQuadrotorControl2024}. Therefore, we
leverage a point-mass model with a velocity Verlet integration scheme,
\begin{align}
  \mathbf{p}_{k+1} &= \mathbf{p}_k + \mathbf{v}_k \Delta t + \frac{1}{2} \mathbf{a}_k (\Delta t)^2\\
  \mathbf{v}_{k+1} &= \mathbf{v}_k + \frac{\mathbf{a}_k + \mathbf{a}_{k+1}}{2} \Delta t\\
  \mathbf{a}_{k+1} &= \mathbf{t}_{k+1} - [0, 0, g]^{\top},
\end{align}
where $\mathbf{p}_k \in \mathbb{R}^3$, $\mathbf{v}_k \in \mathbb{R}^3$,
and $\mathbf{a}_k \in \mathbb{R}^3$ denote the position, velocity, and acceleration of the point
mass at time step $k$, respectively. The time step, $\Delta t$, is fixed during
integration. The constant $g$ denotes gravity.  Since point-mass dynamics do not
maintain an explicit orientation, the orientation is determined uniquely by the
simulator at every time step. The basis vectors for the body frame are defined
such that the body $z$-axis points along the predicted thrust vector
$\mathbf{t}_k$ and the $x$-axis is aligned with the predicted yaw angle $\psi_k$.

\subsection{Network Architecture\label{ssec:network_arch}}

The network takes as input a depth image, a target velocity, goal
importance value, and the robot's state (see~\cref{fig:network_architecture}).
Each input is processed by a dedicated feature extractor and projected into a
192-dimensional vector. These vectors are layer-normalized and summed to form
the input to a gated recurrent unit (GRU)~\citep{cho2014learning} cell.
The GRU combines this input with
the previous hidden state to produce a new hidden state, which encodes relevant
features from both current and past observations.

From this latent representation, a linear layer predicts the desired
mass-normalized thrust and yaw angle. By maintaining
a memory of past observations and actions, the latent representation enables the
policy to generate smooth and consistent control outputs.

The state consists of the current velocity and orientation of the robot. The
target is defined by a desired velocity vector towards the goal and a
goal importance value, defined as the reciprocal of the goal distance. This
importance value provides contextual information about the goal's proximity
relative to obstacles in the depth image, helping the network distinguish
whether the goal lies in front of or behind an obstacle. All input vectors are expressed with respect to
the starting frame of the robot. Our network differs from that
of~\citep{zhang2025learning} in that we use a separate target extractor which
includes a goal importance value, we use normalization layers to balance features, and
we predict a yaw angle in addition to mass-normalized thrust.

\subsection{Loss Functions\label{ssec:loss_functions}}

This section describes the individual loss (or reward) functions
applied at each time step. Together, these terms guide the training of
a stable and effective navigation
policy. The loss functions developed by~\citet{zhang2025learning}
are restated for the sake of completeness.
The losses detailed in~\cref{loss:yaw_prediction,loss:geodesic_gradient}
as well as new terms are noted as novel contributions of this work.

\subsubsection{Obstacle Avoidance Loss\label{loss:obstacle}}

\begin{align}
  L_{\text{clearance}} & = \frac{1}{T} \sum_{k=1}^{T} \beta_1 \ln (1+e^{\beta_2 (d_k - r)}) \label{eq:loss_clearance}                                  \\
  L_{\text{collision}} & = \frac{1}{T} \sum_{k=1}^{T} \lVert \mathbf{v}_k^c \rVert \max (1-(d_k - r), 0)^2 \label{eq:loss_collision}
\end{align}

$L_{\text{clearance}}$ is a softplus penalty on the distance $d_k$ to the nearest
obstacle, encouraging the robot to maintain a safe clearance. The scalar $r$ is the
robot radius. $L_{\text{collision}}$ penalizes the magnitude
of the component of velocity $\mathbf{v}_k^c$ directed toward the closest obstacle.
These losses train the policy to keep a safe distance and reduce
forward velocity when approaching obstacles. These terms are restated
from~\cite{zhang2025learning}.

\subsubsection{Smoothness Loss\label{loss:smoothness}}

\begin{align}
  L_{\text{acc}}  & = \frac{1}{T} \sum_{k=1}^{T} \lVert \mathbf{a}_k \rVert^2 \label{eq:loss_acc}                                \\
  L_{\text{jerk}} & = \frac{1}{T-1} \sum_{k=1}^{T-1} \left\lVert \frac{\mathbf{a}_{k+1} - \mathbf{a}_k}{\Delta t} \right\rVert^2 \label{eq:loss_jerk} \\
  L_{\omega}      & = \frac{1}{T} \sum_{k=1}^{T} \lVert \boldsymbol{\omega} \rVert^2 \label{eq:loss_omega}
\end{align}

Since the policy only predicts a single action at a time, it is important that
the sequence of actions over time form a smooth trajectory that can be executed
safely.  The loss function includes penalties on the $\ell^2$-norm of linear
acceleration and jerk in the inertial frame, as well as angular velocity in the
body frame.  Specifically, $L_{\text{acc}}$ encourages stability near hover at
the target (see \cref{ssec:model_error_results}), while $L_{\omega}$ reduces
abrupt yaw changes by discouraging large angular accelerations.
\Cref{eq:loss_acc,eq:loss_jerk} are restated from~\cite{zhang2025learning},
while~\cref{eq:loss_omega} is a new penalty term where $\boldsymbol{\omega}$ is
computed via the Euler angle Jacobian (see
\cref{eq:time_derivative_mapping_matrix}).

\subsubsection{Target Velocity Loss\label{loss:target_velocity}}
\begin{align}
  L_v             & = \frac{1}{T} \sum_{k=1}^{T} \text{Smooth L1}\left( \lVert \mathbf{v}_k^{\text{set}} - \bar{\mathbf{v}}_k \rVert,0 \right) \label{eq:loss_v} \\
  L_{\text{vmax}} & = \frac{1}{T} \sum_{k=1}^{T} \max(\lVert \mathbf{v}_k \rVert - v_{\max}, 0)^2 \label{eq:loss_vmax}
\end{align}

\Cref{eq:loss_v} encourages the policy to track the target velocity
$\mathbf{v}_k^{\text{set}}$. To allow flexibility for obstacle avoidance while maintaining
long-term progress, we use a \SI{2}{\second} moving average of the actual velocity,
$\bar{\mathbf{v}}_k$, in the loss. This smooths short-term deviations and helps the policy
stay aligned with the overall velocity objective. Consistent with findings in
\citet{zhang2025learning}, this averaging also reduces high-frequency velocity
oscillations during rollouts under full rigid-body dynamics.  Additionally, we
introduce a new penalty term $L_{\text{vmax}}$ for speeds exceeding the maximum
target speed $v_{\max}$, which helps bound the predicted thrusts and prevents
policy divergence.

\subsubsection{Yaw Alignment Loss\label{loss:yaw_prediction}}

\begin{equation}
  L_{\text{yaw}} = - \frac{1}{T} \sum_{k=1}^{T} \mathbf{x}^B_k \cdot \tilde{\mathbf{v}}_k
\end{equation}

The yaw alignment loss $L_{\text{yaw}}$ is defined as the negative inner product
between the body $x$-axis $\mathbf{x}^B_k$ and the exponentially weighted moving
average of the velocity $\tilde{\mathbf{v}}_k$.  This loss term, introduced in
this work, enables the quadrotor to reorient towards its desired direction
of motion to navigate around large obstacles.

\subsubsection{Privileged Information via ToA Map\label{loss:geodesic_gradient}}
\begin{figure}
  \centering
  \subfloat[\label{sfig:geo_dist_map}]{%
    \includegraphics[width=0.49\linewidth]{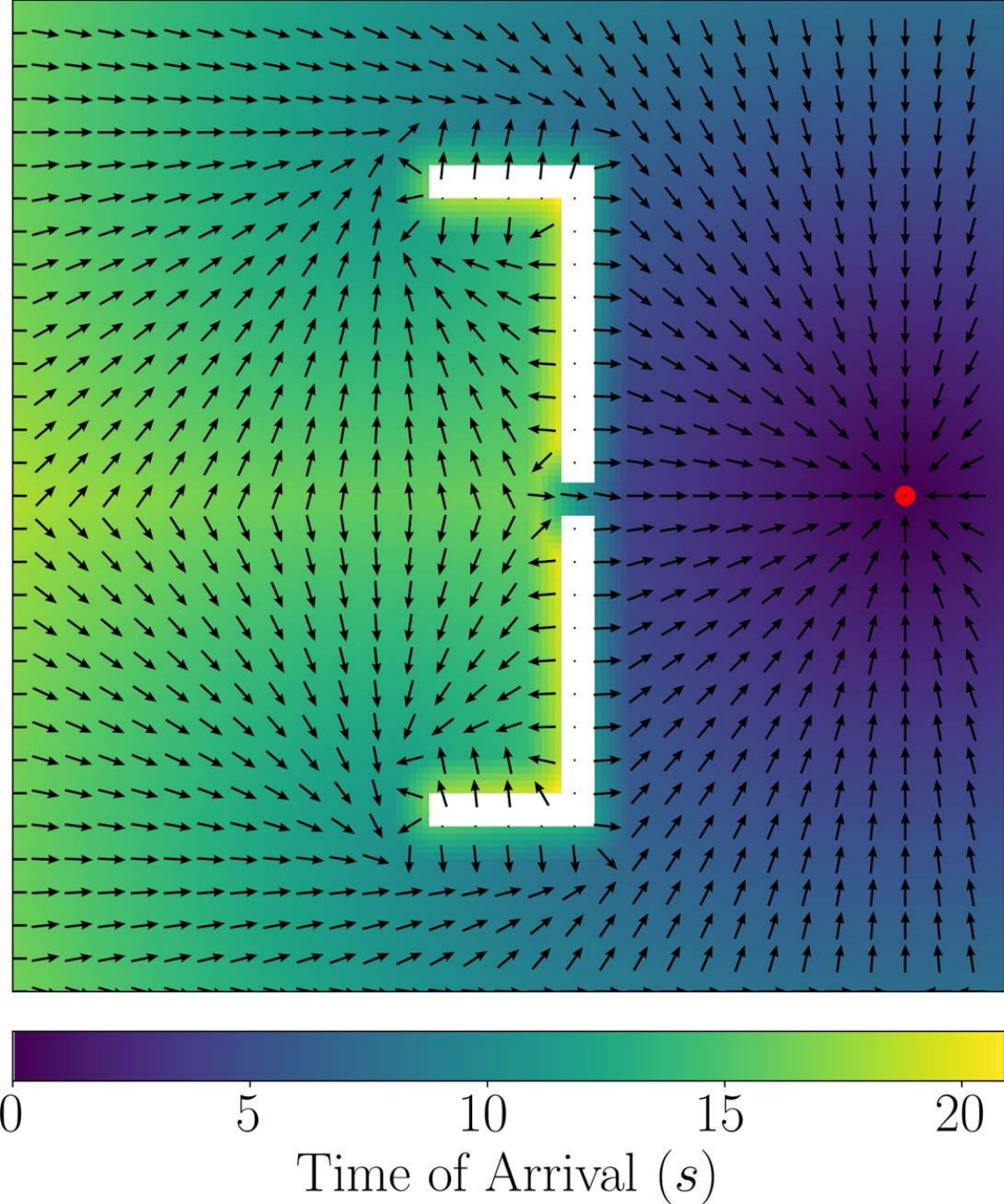}%
  }\hfill
  \subfloat[\label{sfig:shortest_paths}]{%
    \includegraphics[width=0.49\linewidth]{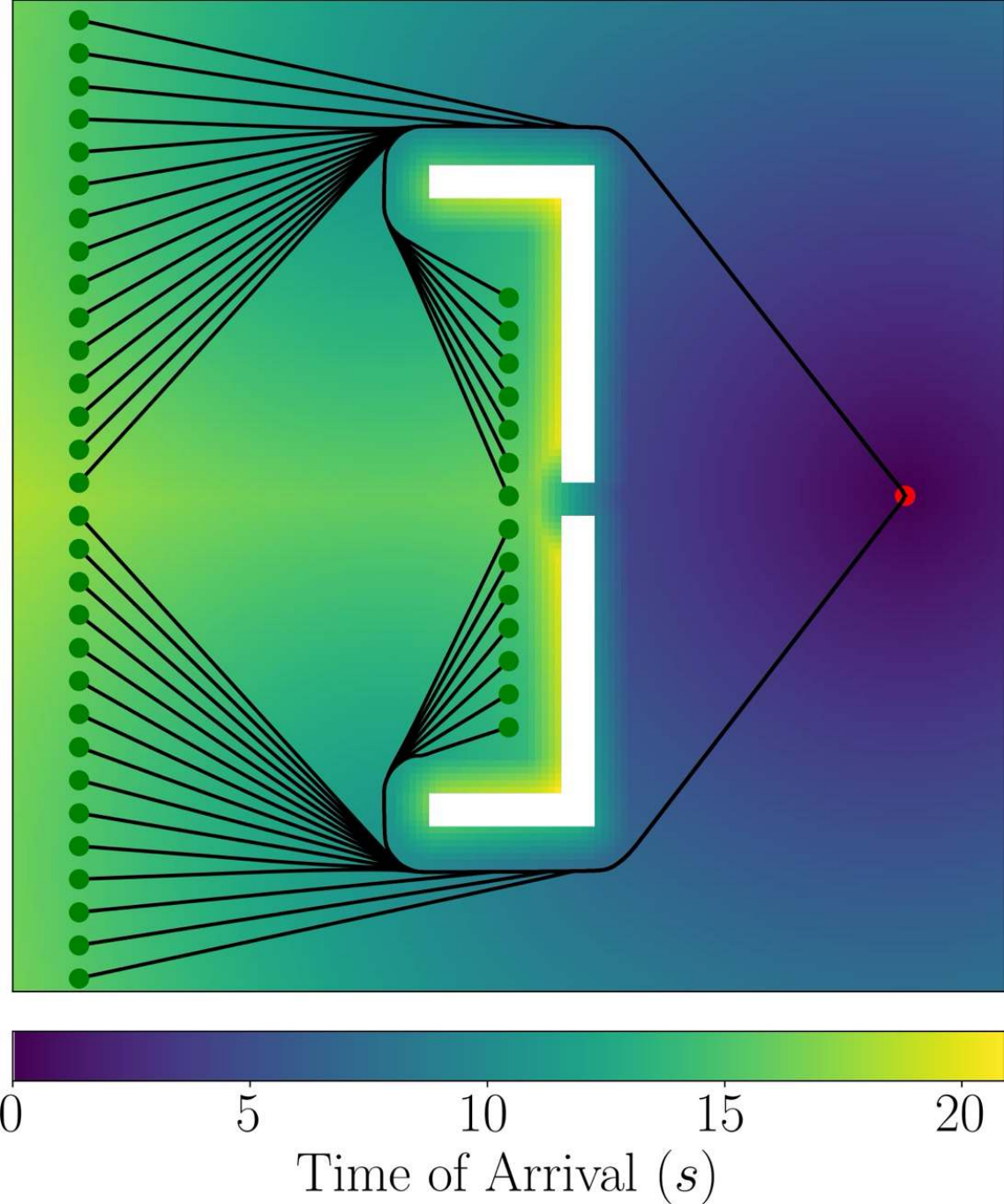}%
  }
  \caption{\protect\subref{sfig:geo_dist_map} Heatmap of time-of-arrival (ToA)
  computed using fast marching method (FMM) and overlayed gradient field.
  \protect\subref{sfig:shortest_paths} Shortest paths along ToA gradient
  from starting points (green dots) to the target (red dot) guides robot
  around concave obstacle regions.
  \label{fig:compute_geodesic}
  }
  \vspace{-0.25cm}
\end{figure}
\red{We observe that existing} Existing obstacle avoidance and collision loss terms in
prior work~\citep{zhang2025learning} are \red{often} insufficient for navigation
around large obstacles. Local collision losses penalize proximity or approach
speed to surfaces but do not actively encourage progress or escape, which can
cause the robot to become stuck in concave obstacle regions.  To address these
limitations, we use time-of-arrival (ToA) maps, whose gradients guide the
robot's velocity toward the goal while avoiding obstacles
(see~\cref{fig:compute_geodesic}).
Through the ToA loss, the policy learns to infer optimal navigation
directions from depth alone, requiring no map at inference.

Let $T(\mathbf{x})$ denote the ToA map, defined as the minimum travel time
from a point $\mathbf{x}$ in free space to the goal position. Travel times are
solutions to the Eikonal equation $|\nabla T(\mathbf{x})| F(\mathbf{x}) = 1$,
which governs how travel time propagates from the goal under a spatially varying
speed $F(\mathbf{x}) > 0$.

While ToA maps naturally account for obstructions due to obstacles, they do
not explicitly consider the risk of being too close to obstacle surfaces, which
can produce unrealistic paths through narrow openings. To address this, we
define $F(\mathbf{x})$ as a piecewise continuous cost function
(\cref{eq:geodesic_velocity}) that reduces wavefront speed near obstacles and
biases the resulting time-optimal paths to maintain safer distances.
\begin{align}
  F(\mathbf{x}) & =
  \begin{cases}
    md(\mathbf{x}) + (v_{\text{slow}} - mr) & \text{if } d(\mathbf{x}) \leq d_{\text{safe}} \\
    1                                       & \text{if } d(\mathbf{x}) > d_{\text{safe}}
  \end{cases} \label{eq:geodesic_velocity} \\
  m    & = \frac{d_{\text{safe}} - v_{\text{slow}}}{d_{\text{safe}} - r},
\end{align}
$r$ is the robot radius, $d(\mathbf{x})$ is the distance to the nearest
obstacle, $d_{\text{safe}}$ is the threshold distance considered safe, and
$v_{\text{slow}}$ is the minimum travel speed near an obstacle,\red{. Points at least
$d_{\text{safe}}$ from obstacles allow the wavefront to travel at \SI{1}{\meter\per\second},
while points closer to obstacles have reduced speed proportional to $d(\mathbf{x})$,
which biases paths away from surfaces (\cref{sfig:shortest_paths})}
biasing paths away from surfaces (\cref{sfig:shortest_paths}).

$T(\mathbf{x})$ is computed on a volumetric grid using the Fast
Marching Method (FMM)~\citep{sethian1996fast} via the \textit{scikit-fmm}
library\footnote{\ifthenelse{\equal{\blindmode}{true}}{\nolinkurl{https://github.com/scikit-fmm/scikit-fmm}}{\url{https://github.com/scikit-fmm/scikit-fmm}}}.
FMM runs in $\mathcal{O}(N \log N)$ time, where $N$ is the number of
grid cells, and each ToA map is pre-computed offline once per training
environment. No ToA map is required at deployment.
Gradients $\nabla T(\mathbf{x})$ are calculated at grid cell centers via finite differencing and
interpolated to provide values in continuous space. The negative gradient then
defines the velocity set-point direction $\mathbf{v}_k^{\text{set}}$ in the
target velocity loss (see~\cref{eq:loss_v}), providing a consistent training signal
that encourages the robot to make progress toward the goal while avoiding
obstacles.

\subsubsection{Total Loss}

\input{tables/hyperparams}
The total loss is a linear combination of all
the loss terms using the coefficients from \cref{tab:hyperparams}:
\begin{align}
  L & = \lambda_{\text{acc}} L_{\text{acc}} + \lambda_{\text{jerk}} L_{\text{jerk}} +\lambda_{\omega} L_{\omega} + \lambda_{v} L_{v} + \lambda_{\text{vmax}} L_{\text{vmax}} \nonumber \\
    & + \lambda_{\text{clearance}} L_{\text{clearance}} + \lambda_{\text{collision}} L_{\text{collision}} + \lambda_{\text{yaw}} L_{\text{yaw}}\label{eq:total_loss}
\end{align}

\subsection{Training\label{ssec:training}}
\begin{figure}
  \centering
  \includegraphics[width=0.49\linewidth,trim=200 0 200 100,clip]{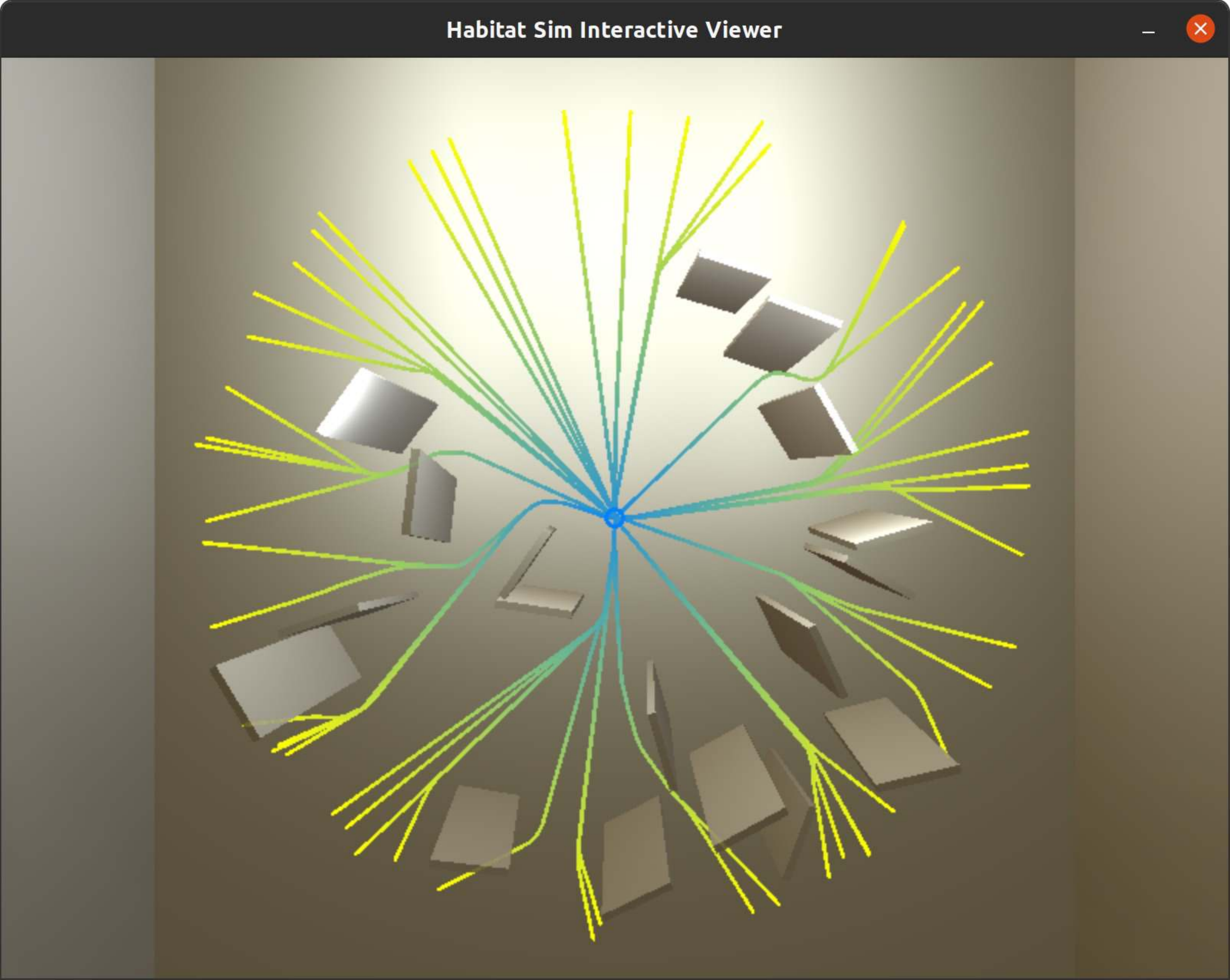}%
  \hfill
  \includegraphics[width=0.49\linewidth,trim=200 0 200 100,clip]{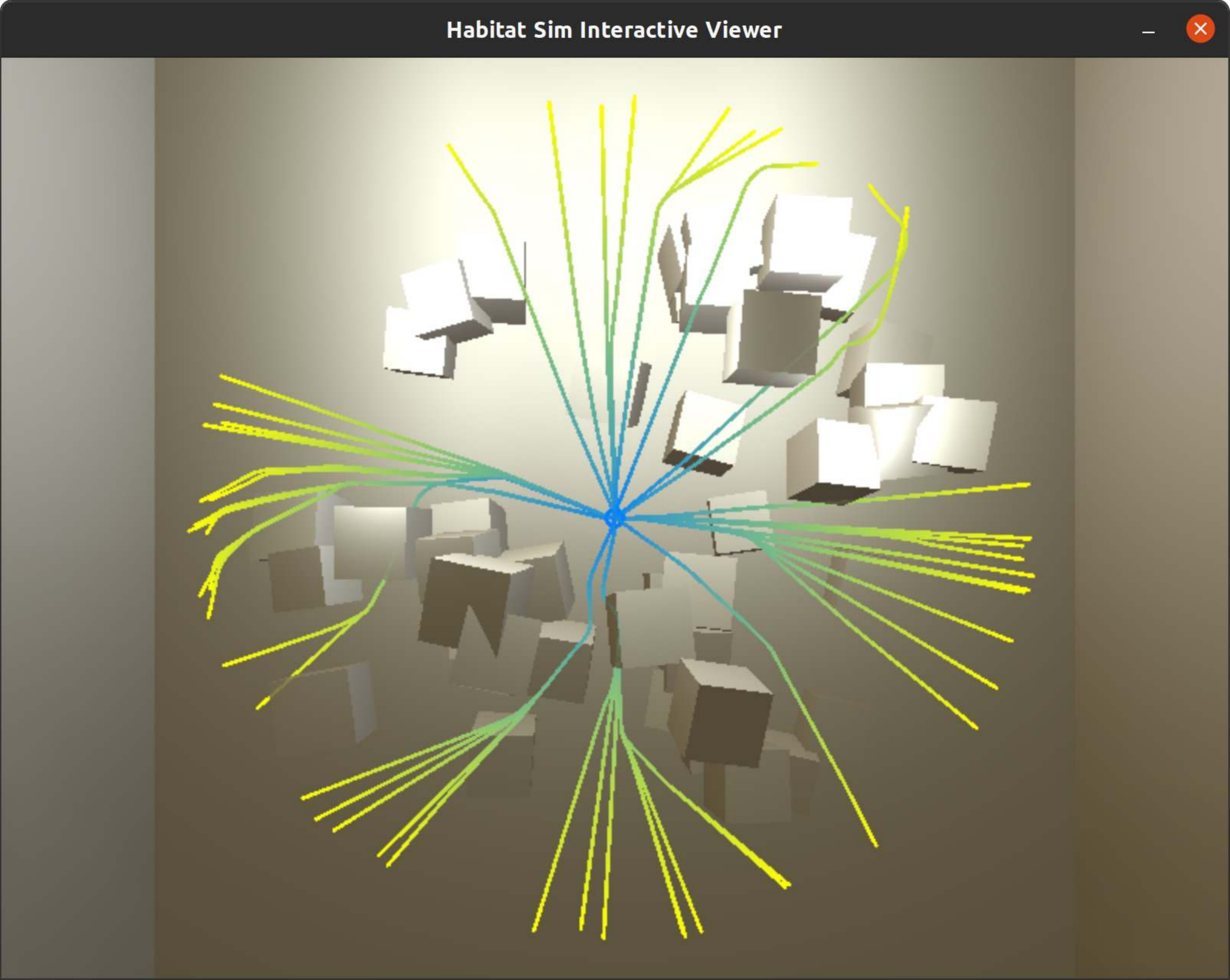}
  \caption{
    \label{fig:habitat-sim-env} Top down view of two cylinder shaped training
    environments with random primitive obstacles and starting points at a fixed
    radius from the goal (blue). Trajectories illustrate paths following the
    time-of-arrival map (yellow to blue).
  }
  \vspace{-0.25cm}
\end{figure}

The policy is trained entirely in simulation with a customized GPU-accelerated
simulator based on VisFly~\citep{liVisFlyEfficientVersatile2024}, which uses
Habitat-Sim~\citep{savvaHabitatPlatformEmbodied2019} to perform rendering and
collision checking in training environments that contain randomly generated
obstacles (i.e., spheres, cylinders, and cuboids; see~\cref{fig:habitat-sim-env}).
A cylindrical shaped training environment enables multiple agents to share a
single environment while observing a variety of ToA gradient directions
simultaneously.  Environments and ToA maps are pre-generated once to reduce
training overhead.
We use back propagation through time (BPTT) and temporal gradient
decay~\citep{zhang2025learning} to accumulate and propagate gradients with a
batch size of 16 over 10K iterations. We find that training with an empty
environment and no collision loss for the first 500 iterations ensures stable
gradients. The policy converges within minutes during this initial stage, while
training in the full environment with the complete loss requires several hours
to achieve proficiency.

\input{tables/domain_randomization}
\begin{figure}
    \centering
    \subfloat[w/o body rate control \label{sfig:without_omega}]{%
        \includegraphics[width=0.99\linewidth]{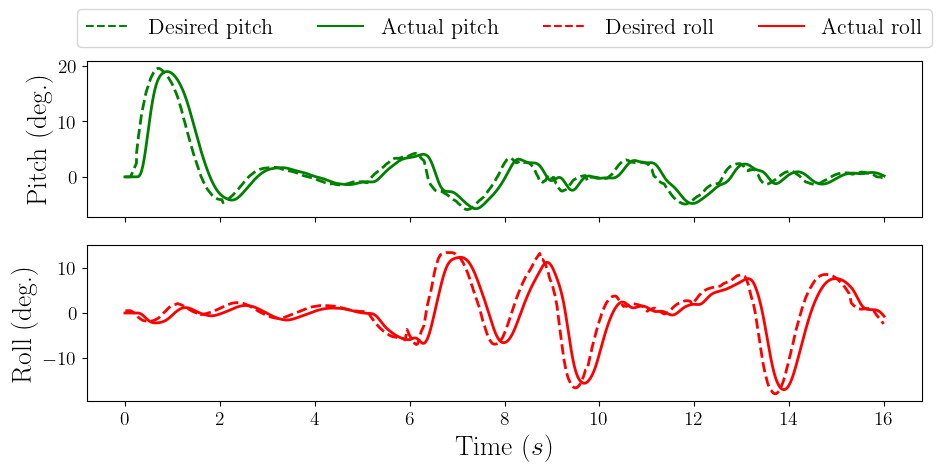}%
    }
    \hfill
    \subfloat[with body rate control\label{sfig:with_omega}]{%
        \includegraphics[width=0.99\linewidth]{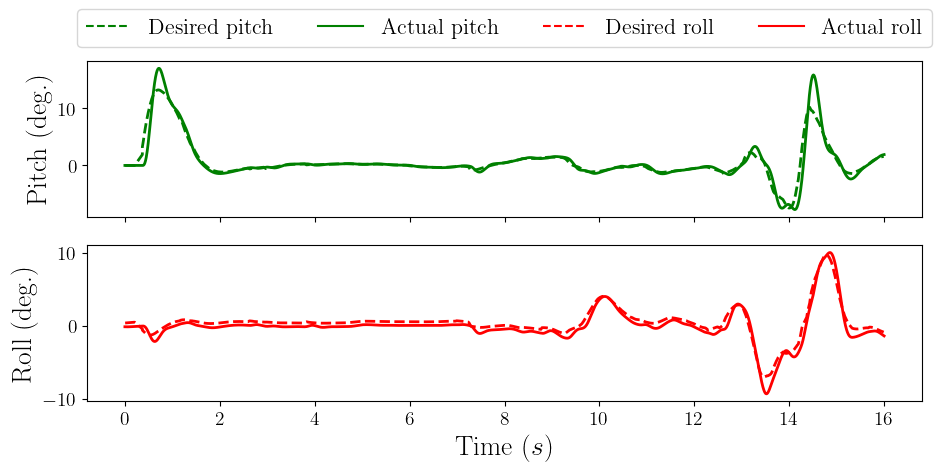}
    }
    \caption{Comparison of attitude control performance without
    (a) and with
    (b) the derivative feedback term,
    $\boldsymbol{\omega}_d$, in the attitude controller.
    \label{fig:attitude_control_performance}}
\end{figure}

\subsection{Body Rate Attitude Control}\label{ssec:body_rate_control}

To bridge the sim-to-real gap between point-mass dynamics in simulation and
rigid-body dynamics in reality, we adopt a PD attitude
controller~\citep{spitzer2021dynamical} that tracks desired thrust $\mathbf{F}$,
orientation $\mathbf{R}_d$, and body rates $\boldsymbol{\omega}_d$.
\red{The desired rotation matrix $\mathbf{R}_d$ is constructed such that the body
$\mathbf{z}^B$ axis aligns with the predicted thrust vector $\mathbf{F}$ and
$\mathbf{x}^B$ aligns with the projection of the desired yaw direction $\psi_d$
onto the plane orthogonal to $\mathbf{z}^B$.}
\red{The desired body rates $\boldsymbol{\omega}_d$ are obtained by first estimating
the Euler angle rates from two consecutive desired attitudes, $\mathbf{R}_{d1}$
and $\mathbf{R}_{d2}$. The relative rotation between them yields the ZYX Euler
angle derivatives $\dot{\phi}, \dot{\theta}, \dot{\psi}$, which are then
converted to body rates via the Euler angle Jacobian:}
The desired rotation $\mathbf{R}_d$ aligns the body $\mathbf{z}^B$ axis with
$\mathbf{F}$ and $\mathbf{x}^B$ with the predicted yaw direction $\psi_d$.
The desired body rates $\boldsymbol{\omega}_d$ are computed from ZYX Euler angle
rates estimated from consecutive desired attitudes via the Euler angle
Jacobian:
\begin{equation}
  \boldsymbol{\omega}_d
  =\begin{bmatrix}
    1 & 0           & -\sin(\theta)           \\
    0 & \cos(\phi)  & \sin(\phi) \cos(\theta) \\
    0 & -\sin(\phi) & \cos(\phi)\cos(\theta)
  \end{bmatrix}
  \begin{bmatrix}
    \dot{\phi} \\ \dot{\theta} \\ \dot{\psi}
  \end{bmatrix} \label{eq:time_derivative_mapping_matrix}
\end{equation}
Including the derivative term $\boldsymbol{\omega}_d$ significantly improves
control response, enabling the attitude controller to achieve the desired
orientation with negligible latency in both simulation and real-world flights.
In contrast, controllers that rely solely on proportional feedback, such as in
\cite{zhang2025learning}, exhibit noticeable control lag, which must be
accounted for during policy training. We show examples contrasting the
control performance in~\cref{ssec:attitude_control_performance}.

\begin{figure*}
  \centering
  \includegraphics[width=0.99\linewidth]{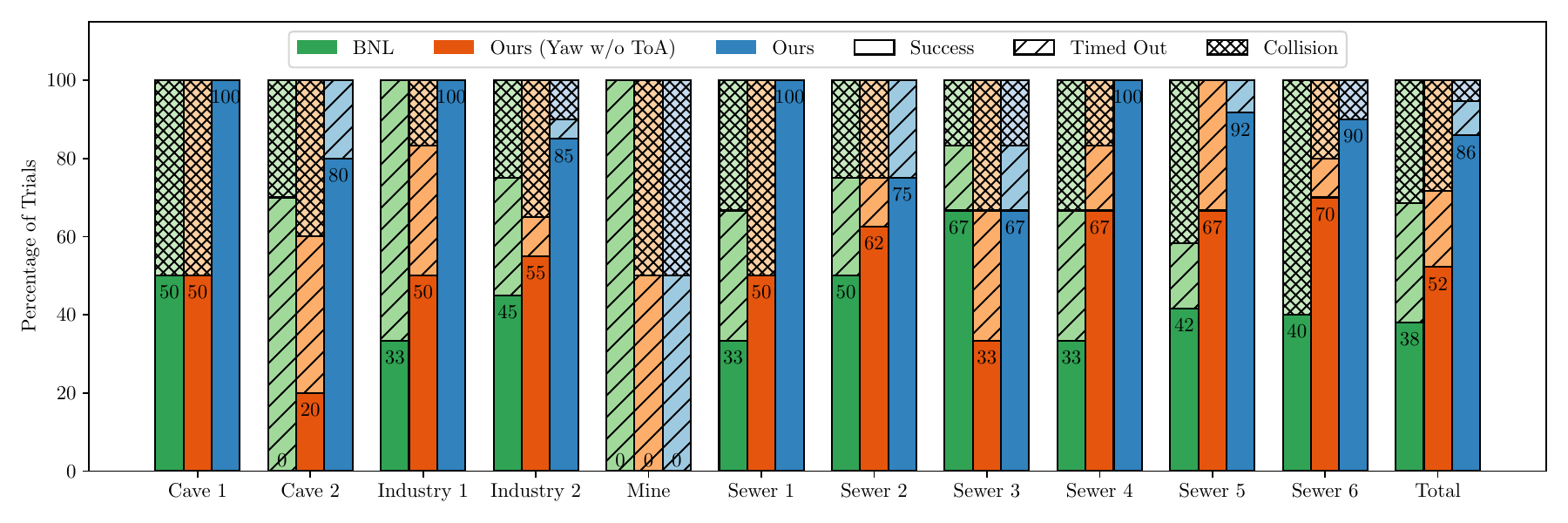}
  \caption{\label{fig:part2_static_success_rate} Planner success rate and
    failure modes across 11 diverse environments. The proposed method
    (\emph{Ours}) achieves the highest success rate and lowest collision rate
    compared to the baseline \citep{zhang2025learning} and the ablated policy
    trained without privileged information (yaw w/o ToA). The \emph{Mine}
    environment features a maze-like corridor which results in poor performance
    from all planners.}
    \subfloat[BNL \xmark]{%
        \includegraphics[width=0.32\textwidth,trim=0 80 1250 300,clip]{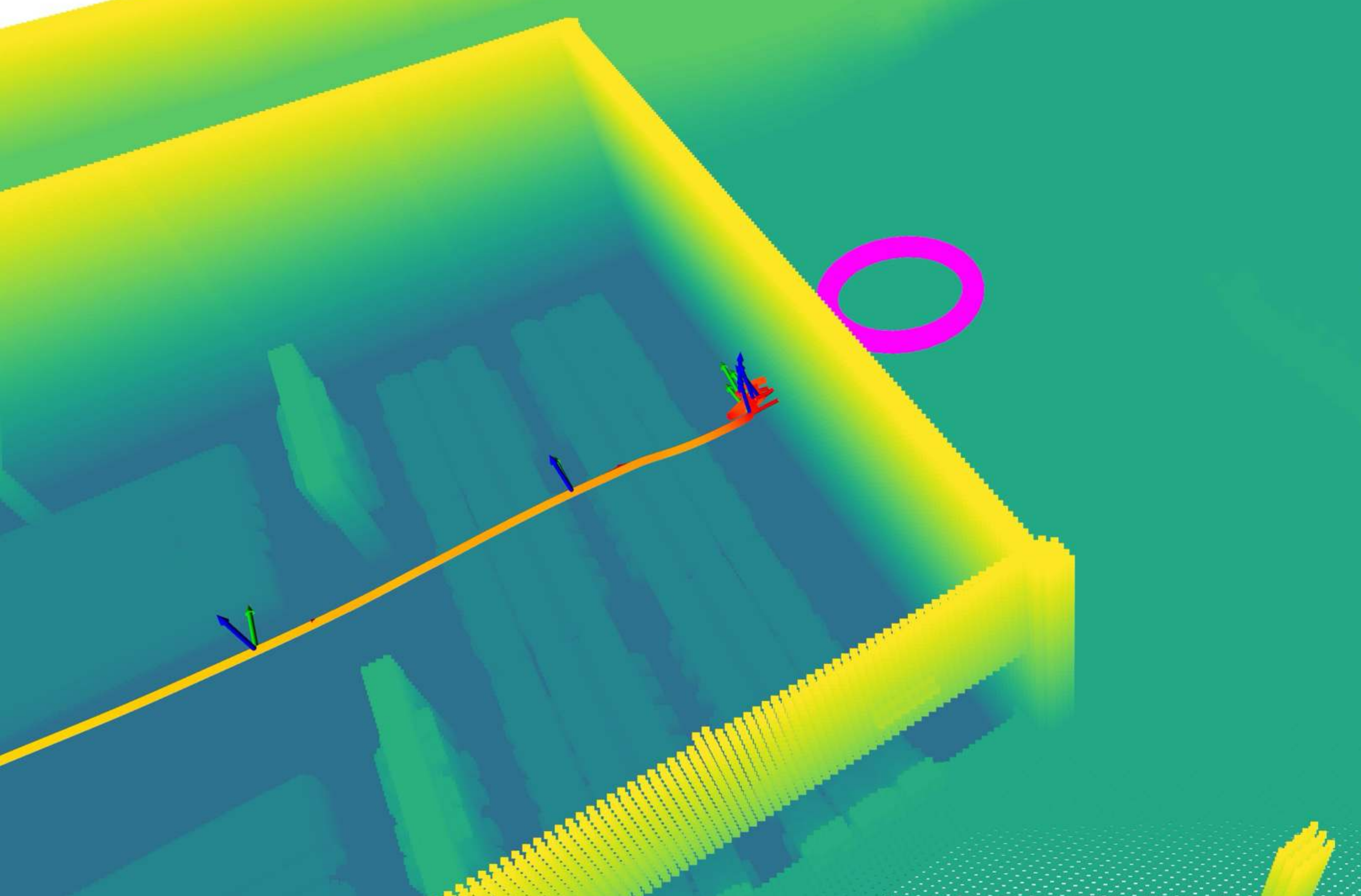}%
    }
    \hfill
    \subfloat[Yaw w/o ToA \xmark]{%
        \includegraphics[width=0.32\textwidth,trim=200 200 1300 350,clip]{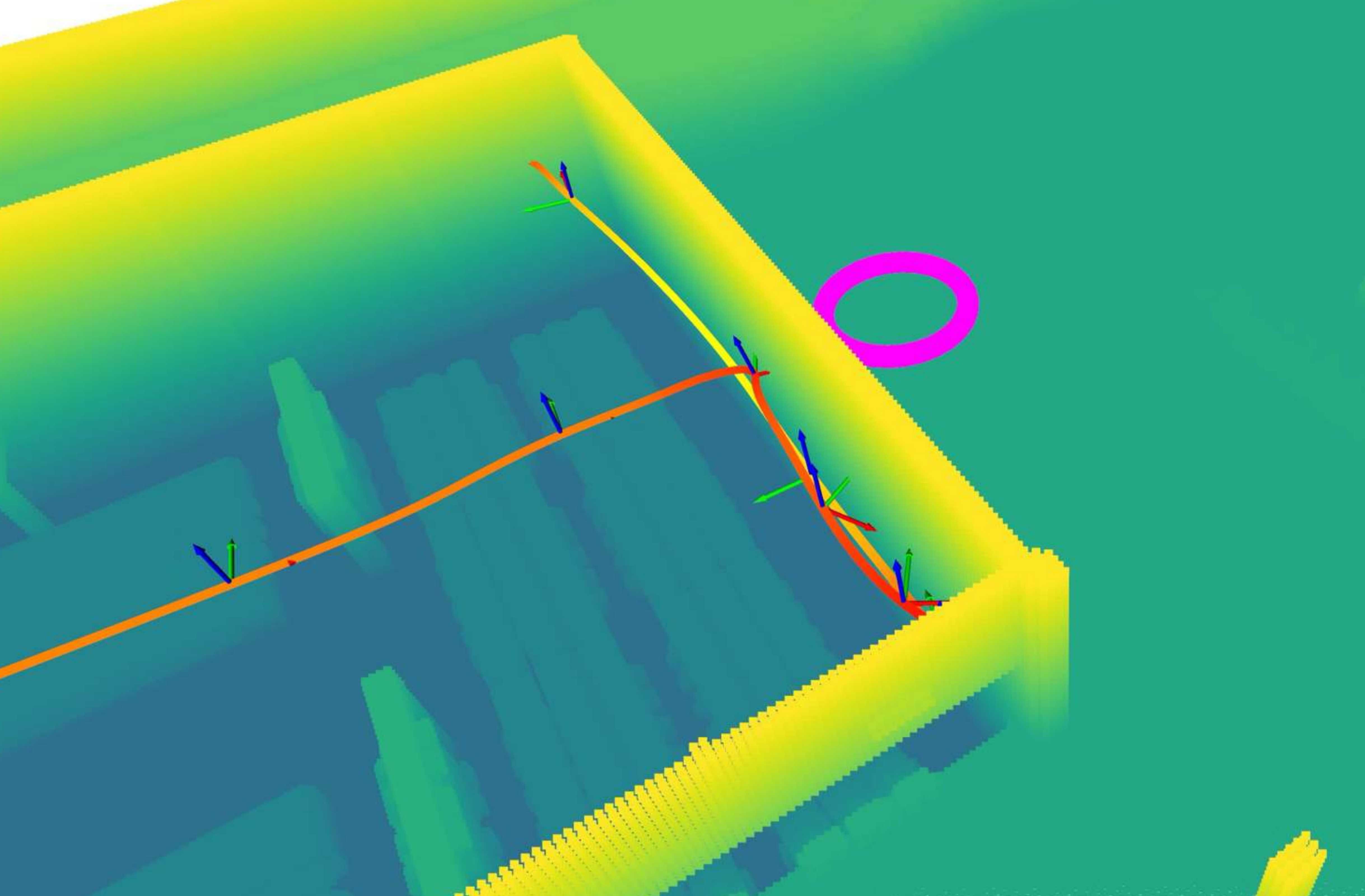}%
    }
    \hfill
    \subfloat[Yaw w ToA \cmark]{%
        \includegraphics[width=0.32\textwidth,trim=200 200 1300 350,clip]{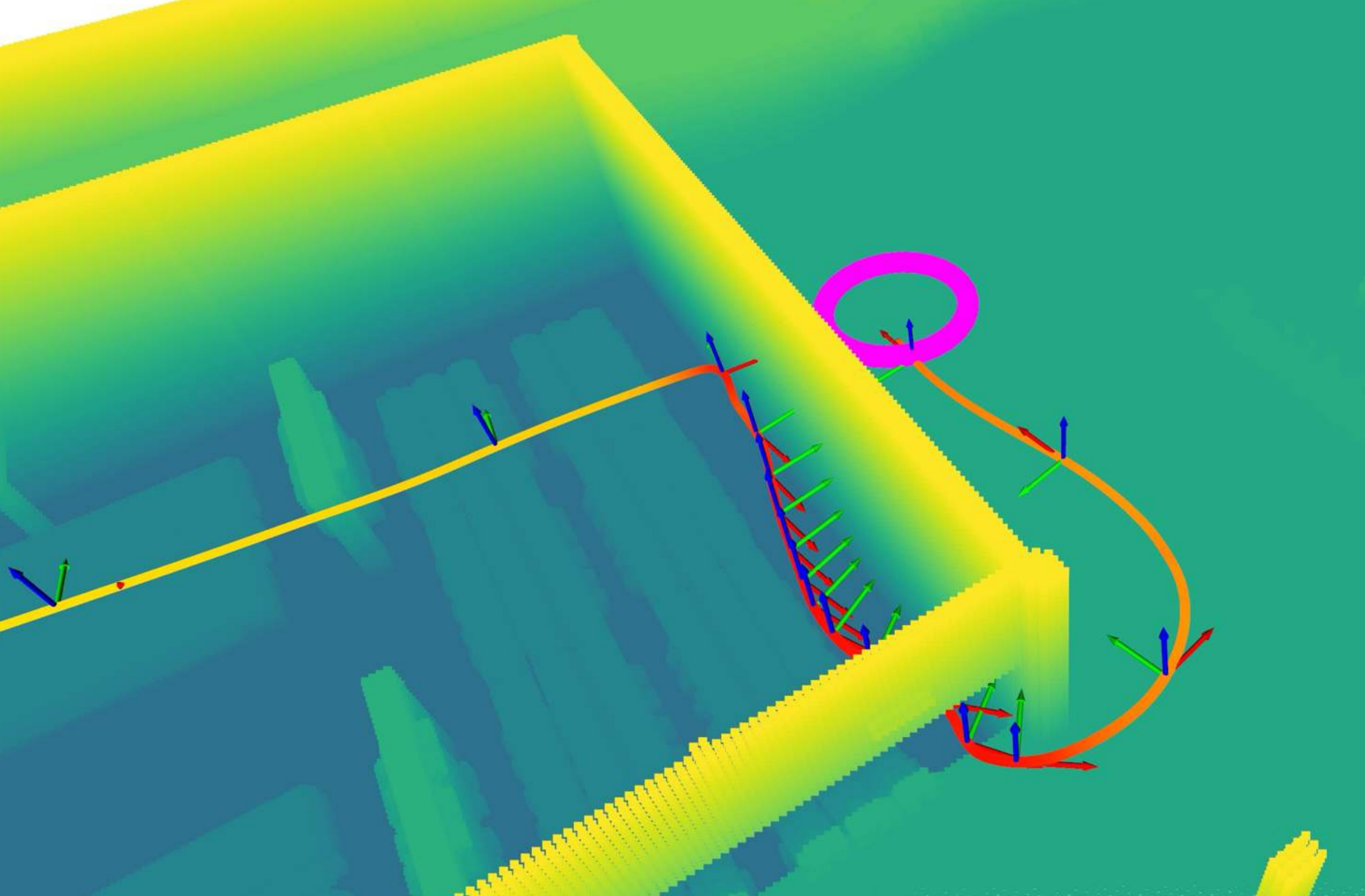}%
    }


    \subfloat[BNL \xmark]{%
        \includegraphics[width=0.32\textwidth,trim=600 380 1500 100,clip]{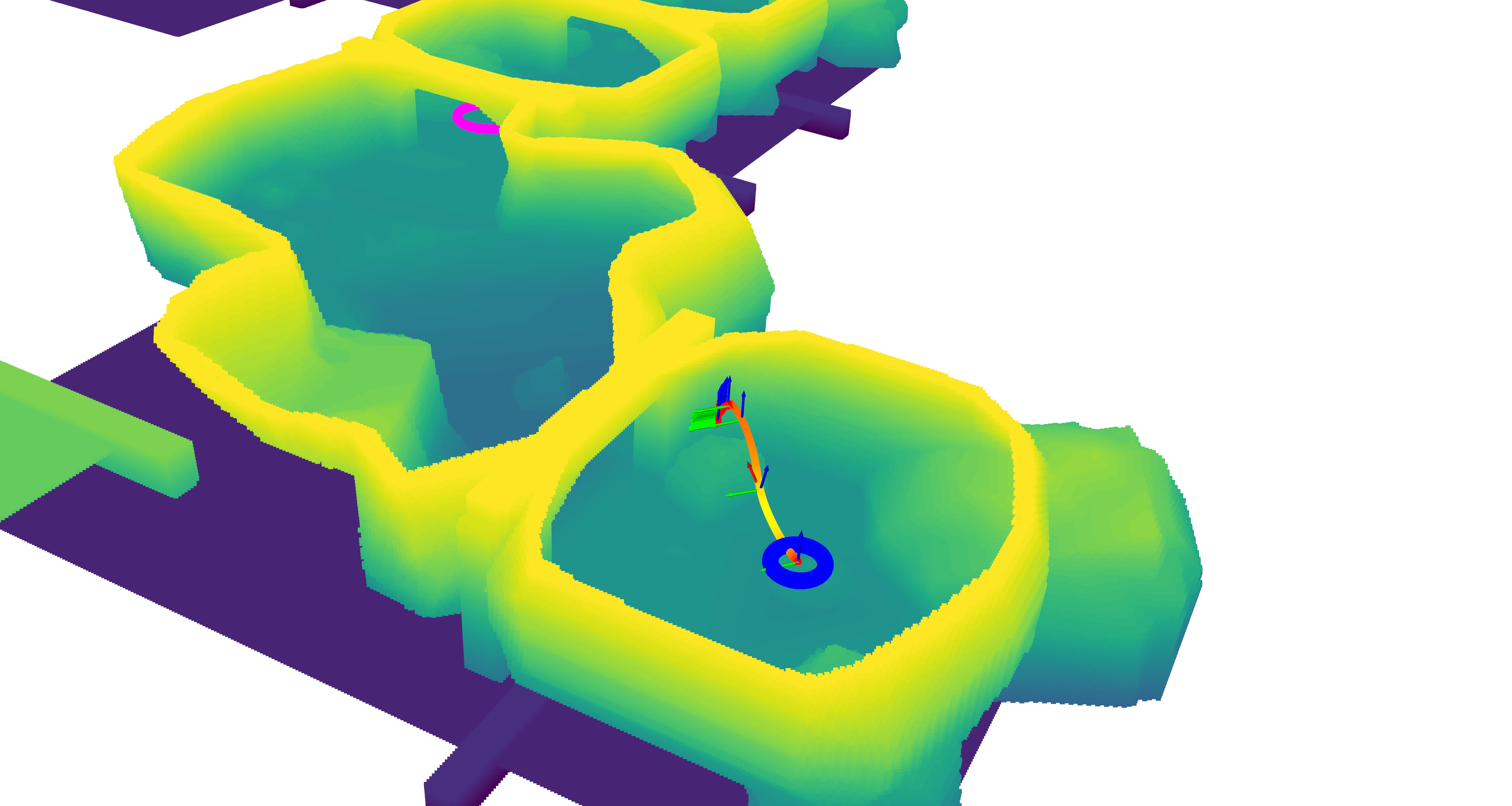}%
    }
    \hfill
    \subfloat[Yaw w/o ToA \cmark]{%
        \includegraphics[width=0.32\textwidth,trim=600 380 1500 100,clip]{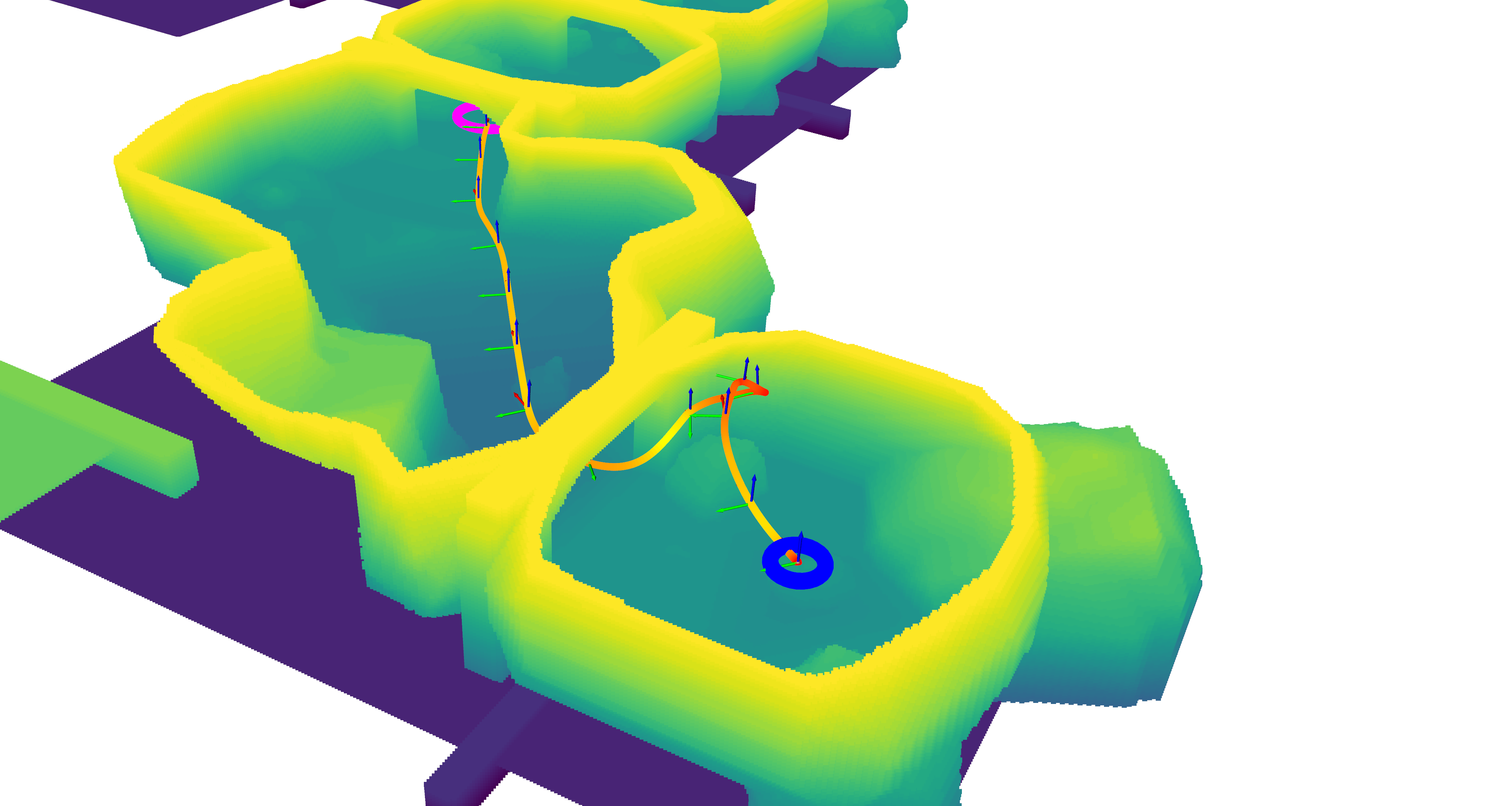}%
    }
    \hfill
    \subfloat[Yaw w ToA \cmark]{%
        \includegraphics[width=0.32\textwidth,trim=600 380 1500 100,clip]{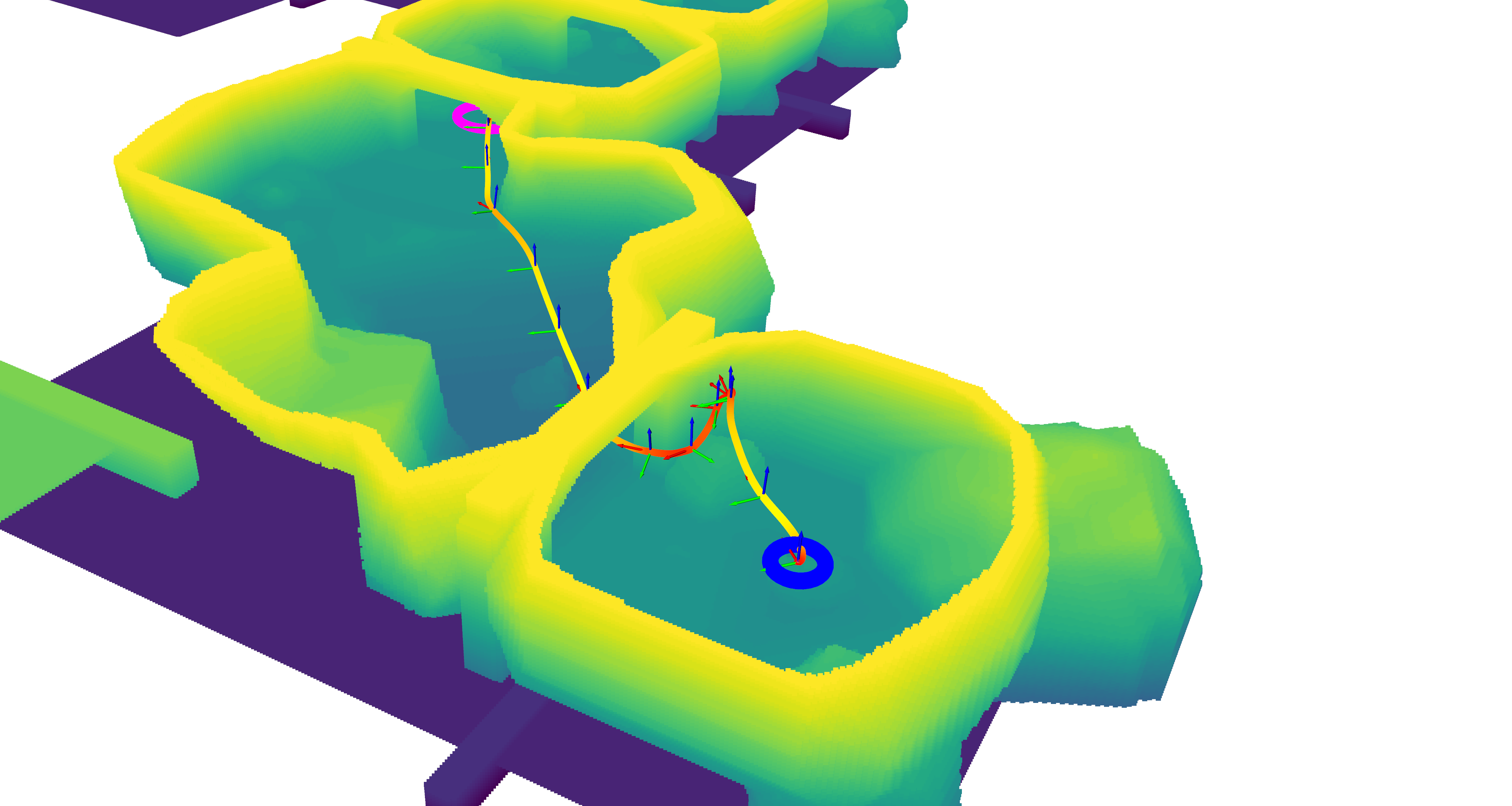}%
    }

    \caption[Qualitative comparison of proposed policy versus ablated policy and
    baseline method. Trajectory profiles overlaid on ground-truth point
    clouds]{\label{fig:sec2_trajectory_qualitative}
      \red{Trajectory profiles} Trajectories overlaid on ground-truth point clouds, with a
    cross-section shown for clarity. \red{Trajectories are colored by speed (red to
    yellow, up to $v_{\text{max}} = 3\ \mathrm{m/s}$), with body frames every 2
    seconds. Start and goal are marked in blue and magenta, respectively.}
    Trajectories are colored by speed (red to yellow, up to
    $v_{\text{max}} = 3\ \mathrm{m/s}$) with body frames every 2\,s.
    Start and goal are marked in blue and magenta.
    Successful trials are marked with a~\cmark~in
    \textit{Industry 2} (top) and \textit{Cave 2} (bottom) environments.
    ToA maps serve as an inductive bias during training only and are
    not available during simulation or hardware evaluation.
    }
    \vspace{-0.25cm}
\end{figure*}

\subsection{Domain Randomization}\label{ssec:domain_randomization}

To ensure platform-agnostic deployment, the policy outputs mass-normalized
thrust (in units of \si{\meter\per\second\squared}), which is later converted to
motor commands using a parametric quadrotor model. Inaccuracies in parameters
such as mass, thrust coefficients, and battery voltage can cause steady-state
errors between expected and actual thrust.

To improve robustness to such discrepancies, we apply domain randomization
during training (\cref{tab:domain_randomization}). We randomize gravity $g$
across rollouts, forcing the policy to adapt its thrust based on velocity
feedback, with smoothness loss discouraging unintended accelerations~(see
experiments in \cref{ssec:model_error_results}). We also randomize initial
position, target velocity, and inject noise into state inputs to simulate sensor
imperfections and improve generalization.

%% file: tables/hyperparams.tex
\setlength{\tabcolsep}{2pt}
\begin{table}
\centering
\caption{Loss function parameters}%
\label{tab:hyperparams}
\setlength{\tabcolsep}{2pt}
\begin{tabular}{llllllllll}
\toprule
$\lambda_{\text{acc}}$ & $\lambda_{\text{jerk}}$ & $\lambda_{\omega}$ & $\lambda_{v}$ & $\lambda_{\text{vmax}}$ & $\lambda_{\text{clearance}}$ & $\lambda_{\text{collision}}$ & $\lambda_{\text{yaw}}$ & $\beta_1$ & $\beta_2$\\
  \hline
0.01 & 0.001 & 0.3 & 4.0 & 1.0 & 6.0 & 6.0 & 1.0 & 2.5 & -6.0 \\
\bottomrule
\end{tabular}
\end{table}

%% file: tables/domain_randomization.tex
\begin{table}
  \small
\centering
\caption{Domain randomization and parameters}. 
\label{tab:domain_randomization}
\begin{tabular}{lll}
\toprule
Value & Distribution & Parameters\\
\midrule
Gravity & $g \sim \mathcal{N}(\mu, \sigma^2)$ & $\mu=9.81$, $\sigma=1.5$\\
Initial height & $z \sim \mathcal{U}(z_{\min}, z_{\max})$ & $z_{\min} = 0.5$\\
& & $z_{\max}= 2.5$\\
Initial $x$ & $x=r_{\text{cyl}} \cos \theta$ & $r_{\text{cyl}}=12.5$\\
Initial $y$ & $y=r_{\text{cyl}} \sin \theta$ & \\
& $\theta \sim \mathcal{U}(0, 2\pi)$ & \\
Target speed  & $v_{\text{target}} \sim \mathcal{U}(v_{\min}, v_{\max})$ & $v_{\min}=1$\\
& & $v_{\max}=5$\\
Velocity noise & $v_{\text{noise}} \sim \mathcal{N}(\mu, \sigma^2) \in \mathcal{R}^3$ & $\mu = 0, \sigma=0.05$\\
Rotation noise & $r_{\text{noise}} \sim \mathcal{N}(\mu, \sigma^2) \in \mathcal{R}^3$ & $\mu = 0, \sigma=0.02$\\
\bottomrule
\multicolumn{3}{l}{\footnotesize 
    $\mathcal{U}$ and $\mathcal{N}$ denote uniform and normal distributions
    respectively.
}\\
\multicolumn{3}{l}{\footnotesize 
    $r_{\text{cyl}}$ denotes the radius of a cylindrical distribution.
}\\
\end{tabular}
\end{table}

%% file: content/simulation_experiments.tex
\subsection{Attitude Control Performance\label{ssec:attitude_control_performance}}

We perform an ablation study comparing the simulated performance between our
attitude control law with body rate control and a controller only tracking
desired orientation as in~\citep{zhang2025learning}. \Cref{sfig:with_omega}
shows that including a derivative term (i.e., the desired body rate,
$\boldsymbol{\omega}_d$) yields negligible control response latency, whereas
\cref{sfig:without_omega} exhibits a latency of approximately
\SI{200}{\milli\second}. Such latency can delay evasive maneuvers, increasing
the risk of collisions in cluttered or dynamic environments.

\subsection{Diverse Simulated Environments}



We evaluate our planner in both simulated indoor and outdoor environments
featuring narrow corridors and large obstacles.  The evaluation uses the 11
out-of-distribution scenarios from~\citep{lee2024rapid},
with depth images rendered by
Flightmare~\citep{song2021flightmare}.
Notably, the policy trained with point-mass dynamics is evaluated
under full rigid-body quadrotor dynamics without access to ToA maps.
1,350 simulation trials are
conducted across all environments and different start-end location pairs.
Further details on the experimental design are available
in~\citep{lee2024rapid}.

We compare our method against two baselines: Back to Newton’s Laws
(BNL)~\citep{zhang2025learning}, trained using the authors’ public
codebase\footnote{
\ifthenelse{\equal{\blindmode}{true}}{\nolinkurl{https://github.com/HenryHuYu/DiffPhysDrone}}{\url{https://github.com/HenryHuYu/DiffPhysDrone}}
}, and an ablated version of our model trained without the ToA privileged information (yaw
w/o ToA). We do not ablate ToA without yaw prediction, as the ToA gradient
may direct the robot toward obstacles outside the depth sensor's field of
view, resulting in unsafe behavior. All approaches are trained using only primitive obstacles and
point-mass dynamics. Performance is measured by success rate and failure modes
(collisions and timeouts), with success defined as reaching the target without
collision within a fixed time limit.

Our method achieves a 86\% success rate, 36\% higher than the baselines, and has
the lowest collision rate across all approaches
(\cref{fig:part2_static_success_rate}). As shown in
\cref{fig:sec2_trajectory_qualitative}, \red{BNL struggles with large, flat obstacles
that require turning or lateral movement, often resulting in collisions.
The ablated model, lacking geometric awareness, frequently gets stuck in cluttered
scenes. In contrast, our full model learns to yaw around large obstacles and
leverage ToA map cues to identify collision-free paths through free space.}
The BNL policy struggles to avoid large, flat
obstacles due to its fixed heading towards the target.  The yaw w/o ToA policy
achieves a higher success rate than BNL in environments requiring significant
reorientation (e.g., Industry~2), confirming that yaw alignment alone enables
the policy to turn around large obstacles; however, without ToA guidance it
frequently times out in concave regions.  Our full model yaws around large
obstacles and learns global navigation cues from the ToA loss, yielding the
highest success rate across all environments.


%% file: content/hardware_experiments.tex
\begin{figure*}
  \subfloat[Policy without gravity randomization\label{fig:no_gravity_compensation}]{
  \begin{minipage}[t]{0.49\linewidth}
    \includegraphics[width=0.245\linewidth]{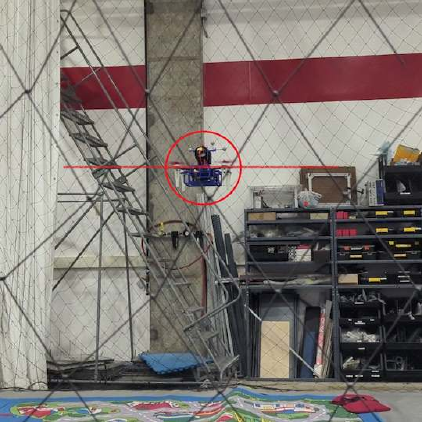}\hfill
    \includegraphics[width=0.245\linewidth]{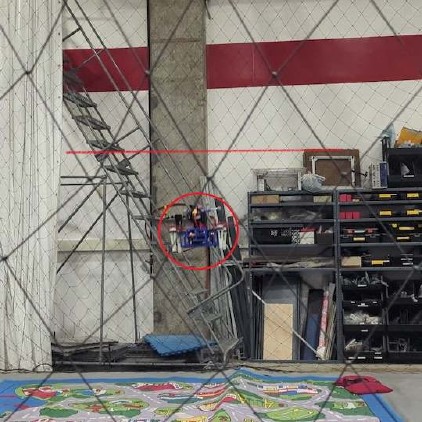}\hfill
    \includegraphics[width=0.245\linewidth]{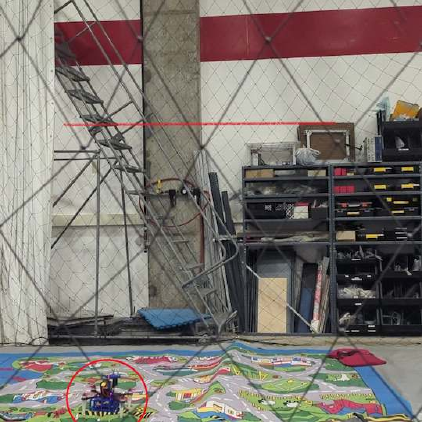}\hfill
    \includegraphics[width=0.245\linewidth]{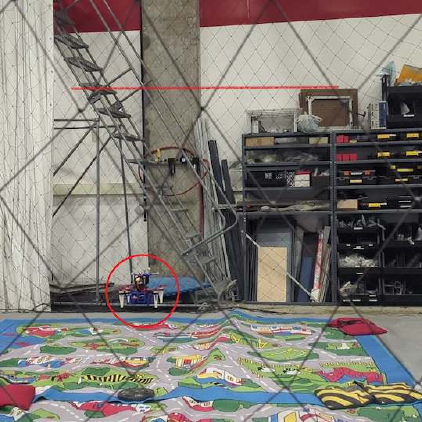}%
    \vspace{0.1cm}
    \includegraphics[width=0.99\linewidth]{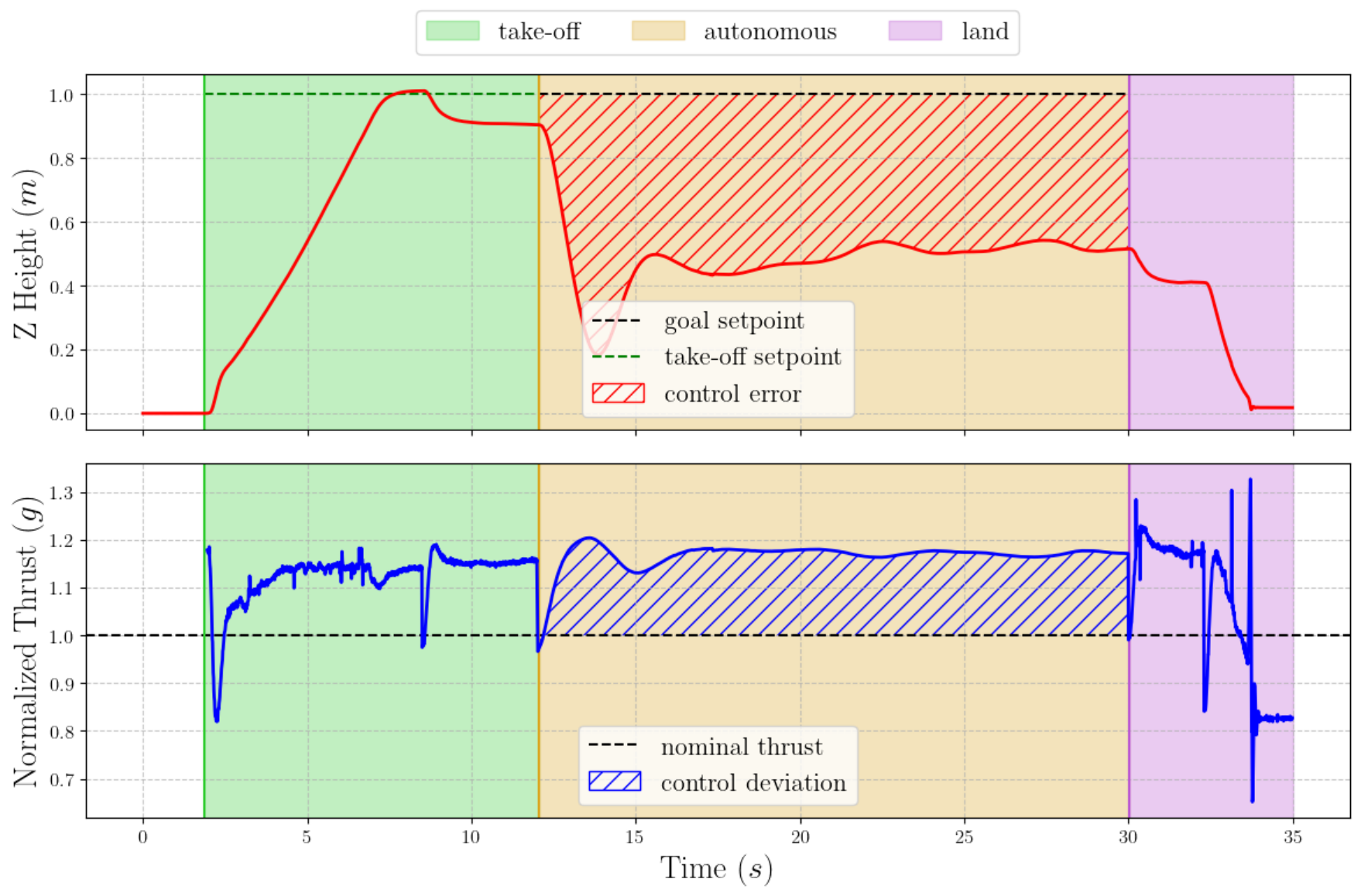}%
    \end{minipage}}\hfill
  \subfloat[Policy with gravity randomization\label{fig:gravity_compensation}]{
    \begin{minipage}[t]{0.49\linewidth}
      \includegraphics[width=0.245\linewidth]{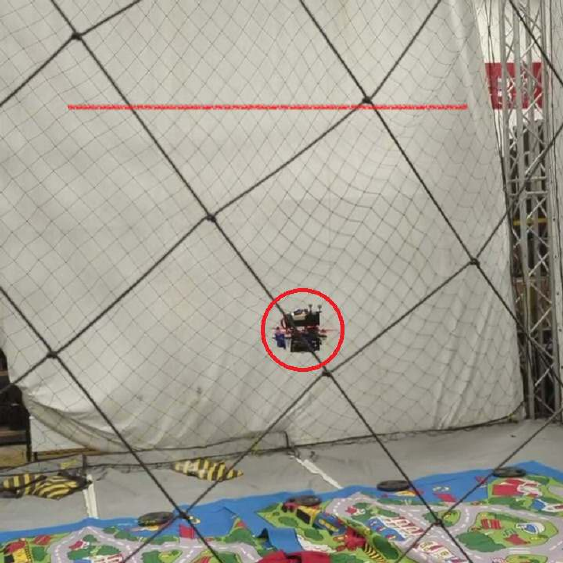}\hfill
      \includegraphics[width=0.245\linewidth]{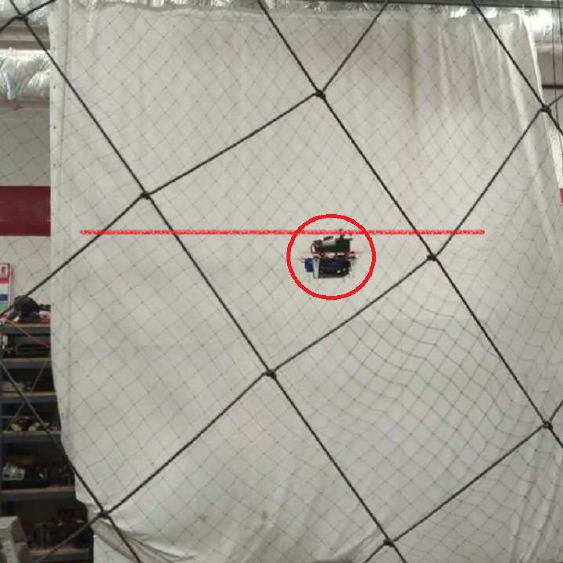}\hfill
      \includegraphics[width=0.245\linewidth]{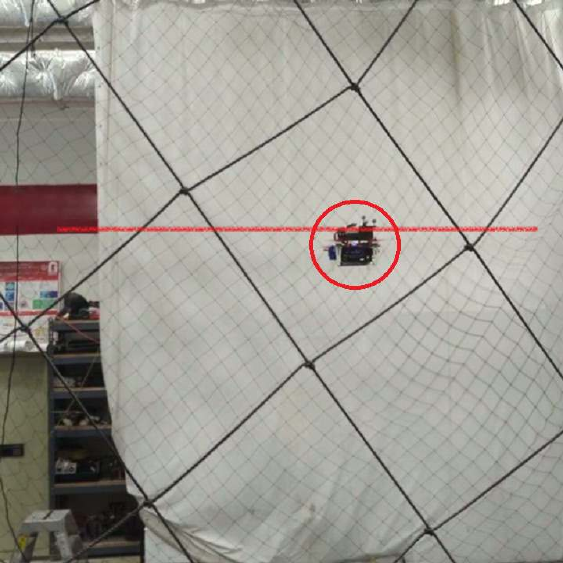}\hfill
      \includegraphics[width=0.245\linewidth]{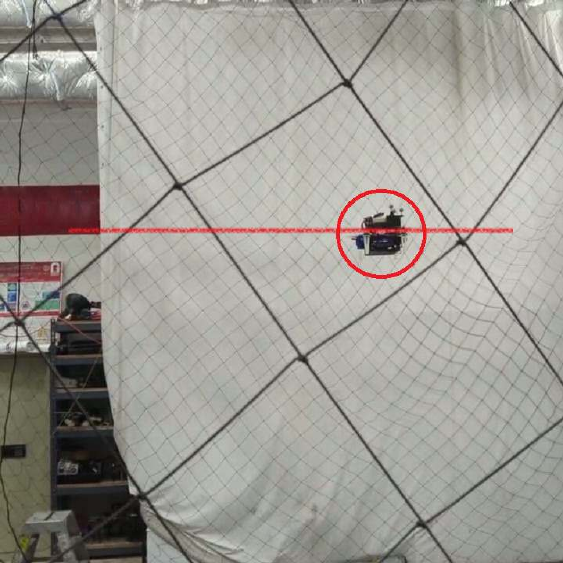}%
      \vspace{0.1cm}
      \includegraphics[width=0.99\linewidth]{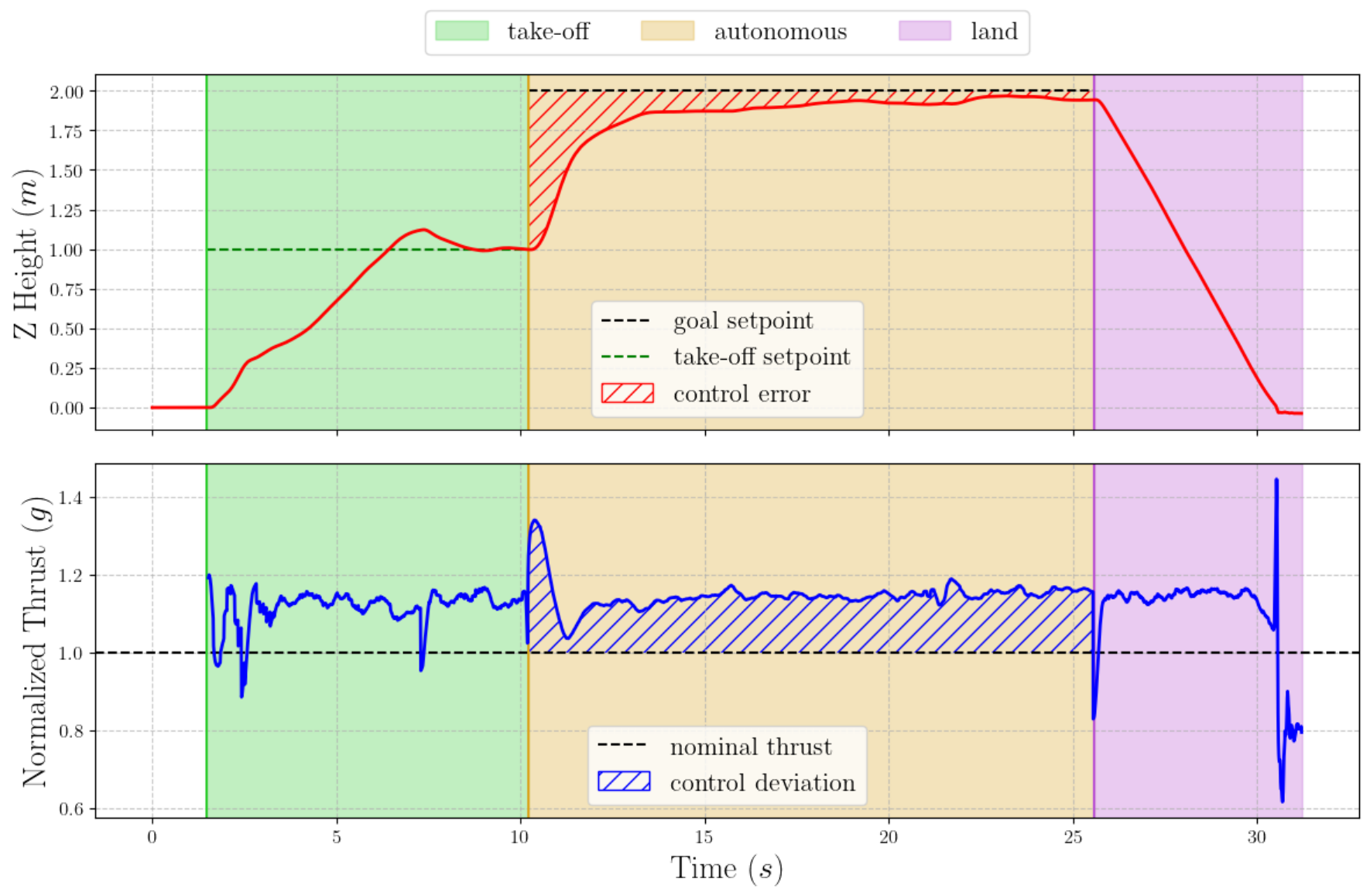}%
    \end{minipage}
    }
    \caption{\label{fig:hardware_ablation_domain_ranodmization}
    \red{Hardware ablation experiments studying the effect of training a policy
    without gravity randomization \protect\subref{fig:no_gravity_compensation}
    and the performance improvement \protect\subref{fig:gravity_compensation}
    when gravity randomization is applied. The top row shows snapshots of the
    quadrotor in flight with the red line indicating the goal setpoint.
    In~\protect\subref{fig:no_gravity_compensation}, the start and goal
    positions coincide, but upon entering autonomous mode, the vehicle quickly
    loses altitude. In contrast, when domain randomization is used
    \protect\subref{fig:gravity_compensation}, the robot gains altitude and
    reaches the goal setpoint. The middle row plots the quadrotor’s z-height
    measured by a motion capture system. With domain randomization, altitude
    error is significantly reduced compared to the non-randomized case. The
    bottom row shows the normalized thrust predicted by the policy, with the
    domain randomized policy producing an initial thrust of $1.3g$, substantially
    higher than in the non-randomized case, enabling better compensation for
    platform modeling inaccuracies.}
    Hardware ablation comparing policies trained
    without~\protect\subref{fig:no_gravity_compensation} and
    with~\protect\subref{fig:gravity_compensation} gravity randomization.  Top
    row: flight snapshots with goal setpoint in red.
    In~\protect\subref{fig:no_gravity_compensation}, the start and goal
    positions coincide, but upon entering autonomous mode, the vehicle quickly
    loses altitude.  With
    randomization~\protect\subref{fig:gravity_compensation}, the robot gains
    altitude and reaches the goal setpoint.  Middle row: plot of z-height from
    motion capture showing significantly reduced altitude error with the gravity
    randomized policy. Bottom row: plot of normalized thrust predicted by the
    policy, where the gravity randomized policy initially outputs $1.3g$,
    substantially higher than the policy without gravity randomization, compensating
    for modeling inaccuracies.
    }
    \vspace{-0.5cm}
\end{figure*}
A custom quadrotor platform was developed to support hardware experimentation.
The vehicle spans \SI{15}{\centi\meter} from rotor to rotor and is equipped with a
forward-facing Intel RealSense D456 depth camera, a downward-facing Lightware SF20/C range
finder, and a Matrix Vision BlueFox2 global shutter greyscale camera. All onboard
computation is performed by an NVIDIA Orin NX module with 16 GB of RAM. The
system has a total mass of \SI{1.7}{\kilogram}.  A TBS Lucid H7 flight
controller runs custom Betaflight firmware\footnote{
\ifthenelse{\equal{\blindmode}{true}}{\nolinkurl{https://github.com/betaflight/betaflight}}{\url{https://github.com/betaflight/betaflight}}
} to provide IMU data at \SI{1000}{\hertz}. For GPS-denied state estimation, we employ a
monocular visual-inertial navigation system developed
by~\citet{Yao-2020-120702}. The RealSense camera generates depth images at \SI{60}{\hertz}
with a resolution of 640 $\times$ 480 pixels and the policy runs at \SI{50}{\hertz}.

\subsection{Domain Randomization Experiments}\label{ssec:model_error_results}

We conduct an ablation study to evaluate whether domain randomization
compensates for modeling errors, comparing policies trained with and without
randomized gravity. A major source of modeling error stems from the
thrust-to-RPM mapping, which is empirically derived under benchtop conditions
that don't account for in-flight factors like battery voltage drop and
frame-induced airflow disturbances. As a result, the quadrotor
exhibits steady-state thrust mismatches; for example, it requires sending
thrust commands of approximately $1.15g$ to hover instead of the expected $1g$
(\cref{fig:gravity_compensation}).

To address this, we train policies with gravity sampled from a normal
distribution (see \cref{ssec:domain_randomization}), encouraging the policy to
learn a closed-loop feedback mechanism. We evaluate both policies in a hover
task where the policy is provided a target velocity vector proportional to the
distance to a fixed goal setpoint.  The non-randomized policy initially outputs
exactly $1g$ of thrust and fails to maintain altitude
(\cref{fig:no_gravity_compensation}), while the randomized policy increases
thrust up to $1.3g$ achieving stable hover at the goal setpoint
(\cref{fig:gravity_compensation}). These results support our hypothesis that
gravity randomization induces adaptive behavior that compensates for thrust
modeling inaccuracies during deployment.


\subsection{Outdoor Navigation Experiments\label{ssec:outdoor_experiments}}
\begin{figure}
  \centering
  \includegraphics[width=0.999\linewidth]{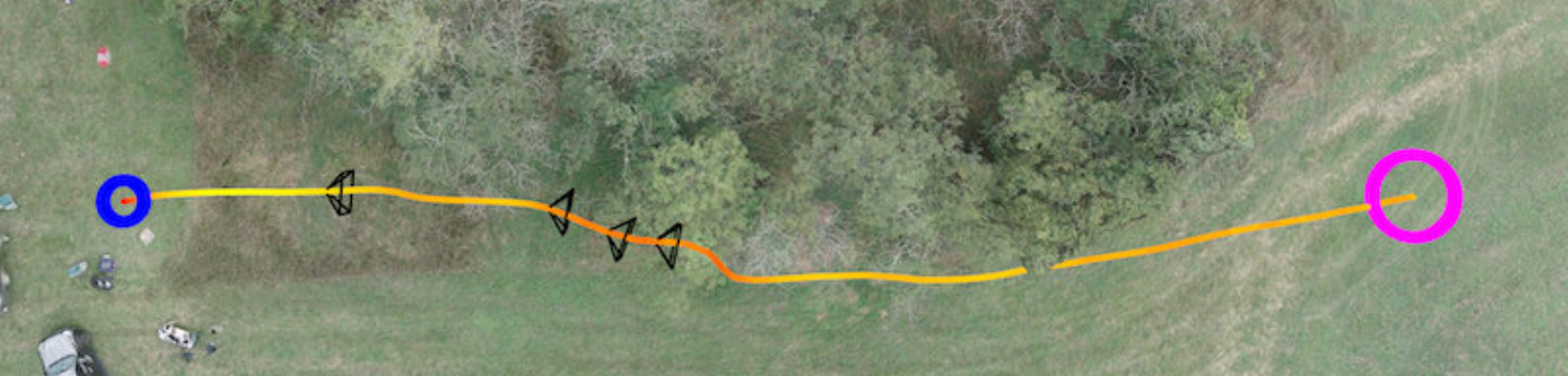}%
  \vspace{0.05cm}
  \includegraphics[width=0.245\linewidth,trim=0 1450 40 0,clip]{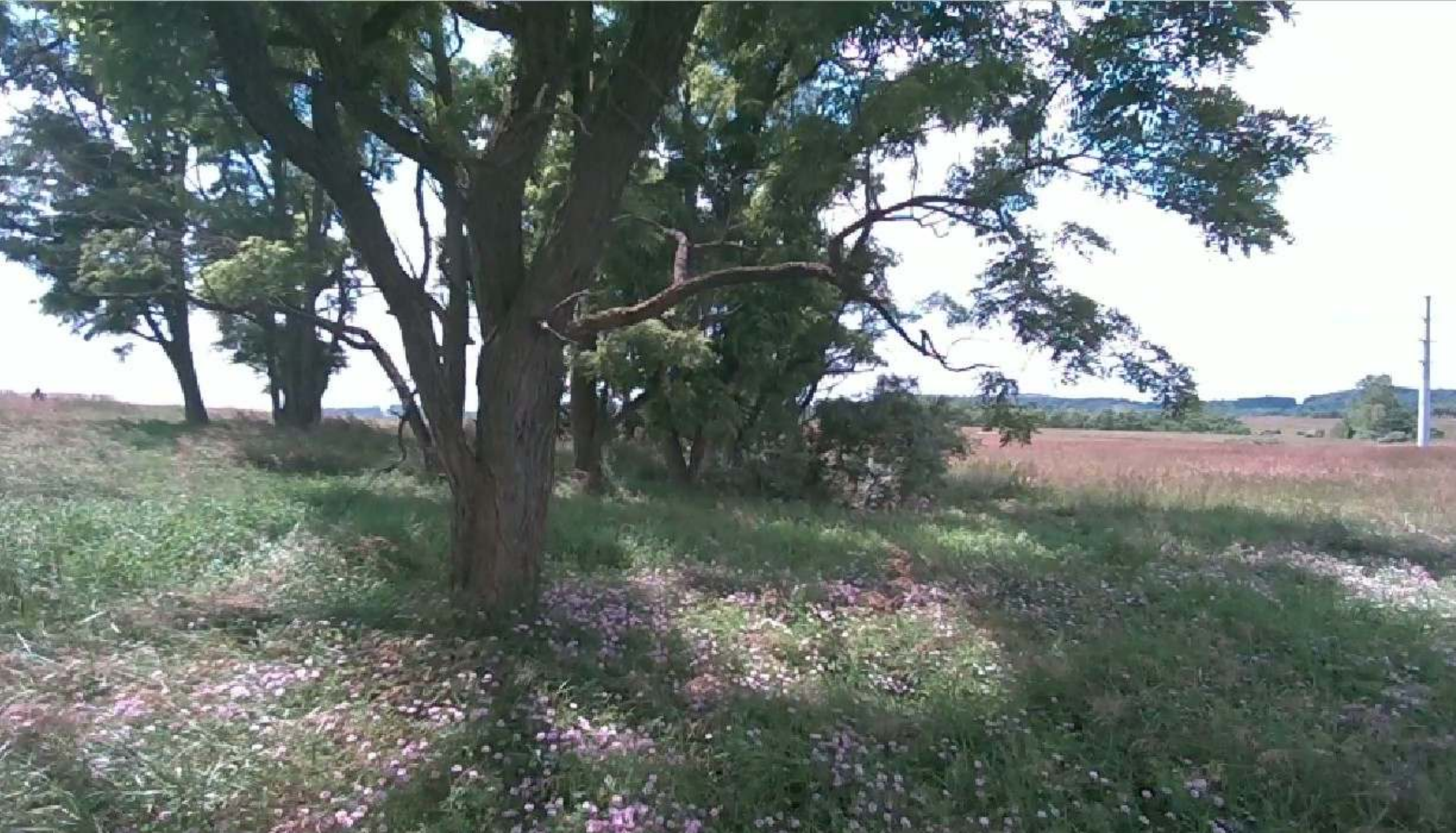}\hfill
  \includegraphics[width=0.245\linewidth,trim=0 1450 40 0,clip]{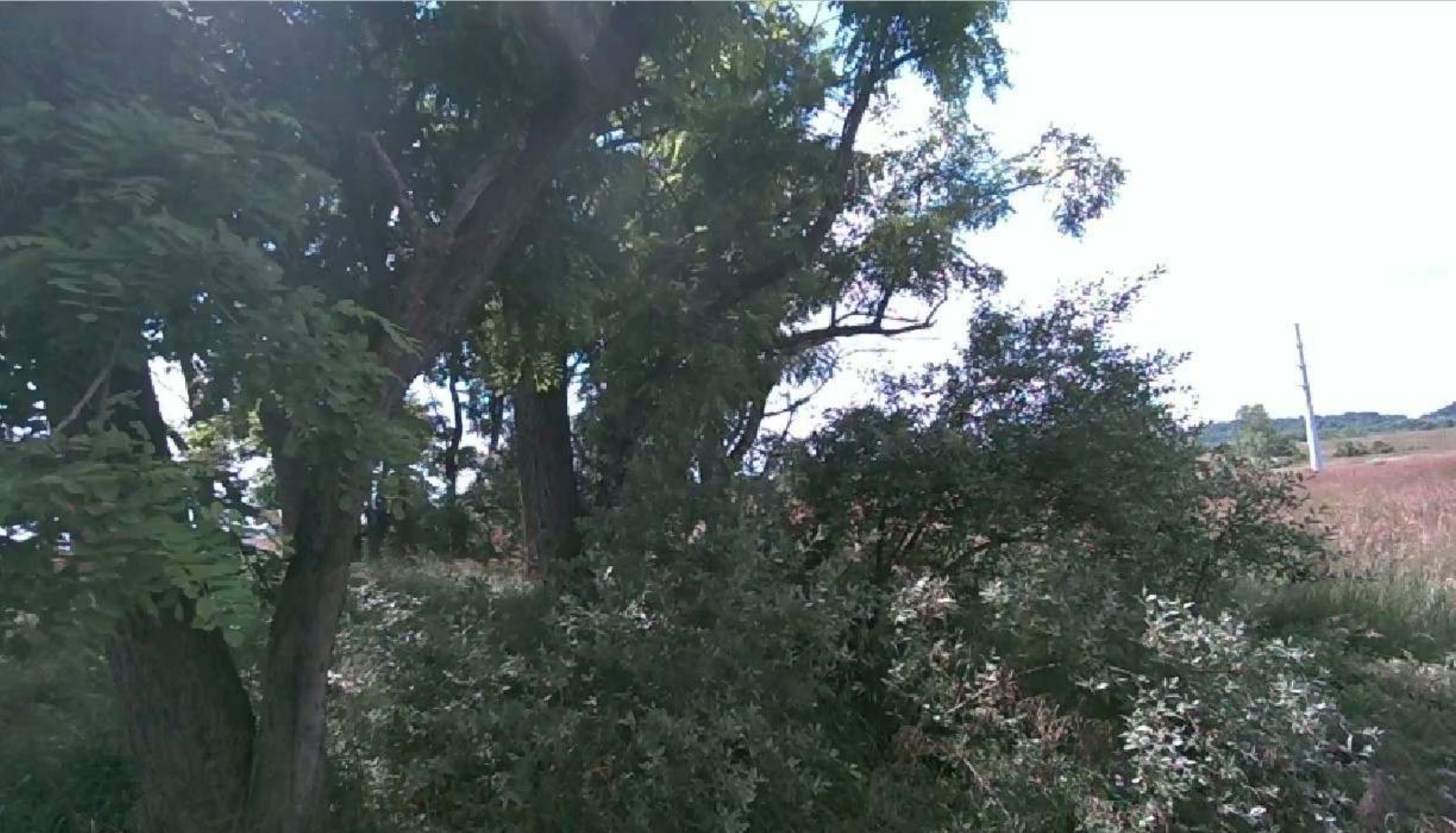}\hfill
  \includegraphics[width=0.245\linewidth,trim=0 1450 40 0,clip]{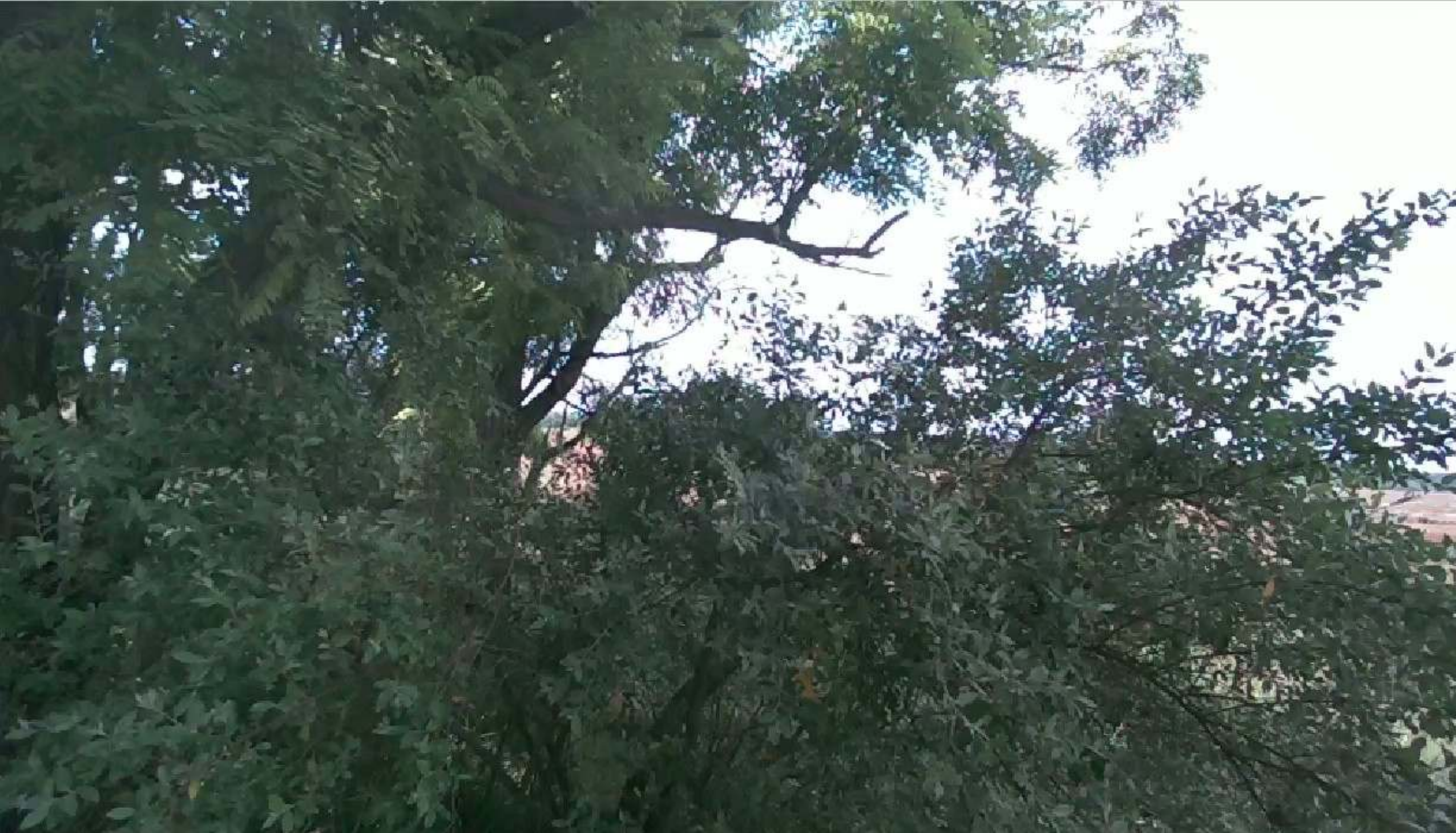}\hfill
  \includegraphics[width=0.245\linewidth,trim=0 1450 40 0,clip]{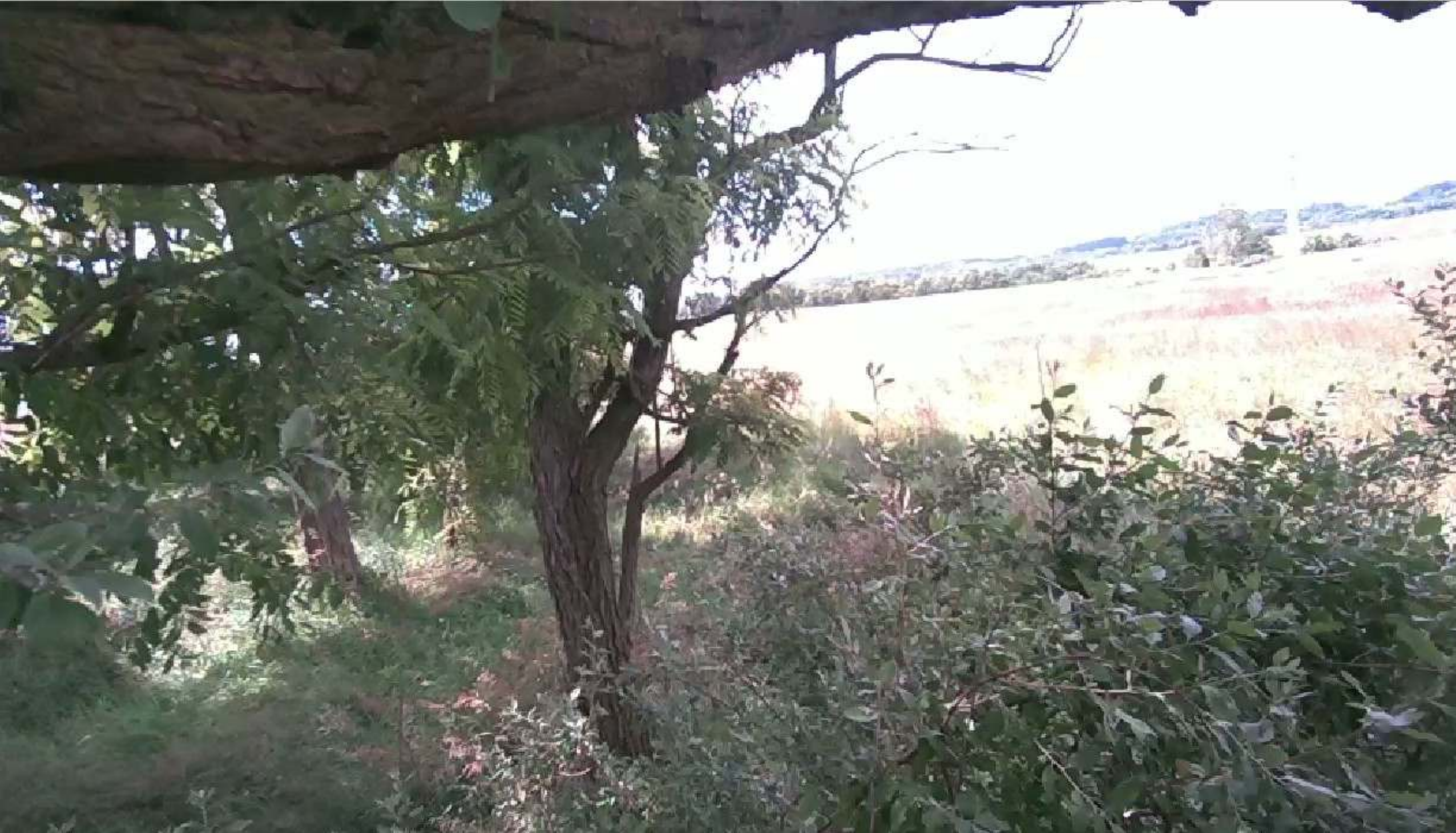}%
  \vspace{0.05cm}
  \includegraphics[width=0.245\linewidth,trim=265 100 290 1503,clip]{figures/hardware/forest/eris02_250702_D-2025-07-22-frame-at-0m21s.pdf}\hfill
  \includegraphics[width=0.245\linewidth,trim=265 100 290 1503,clip]{figures/hardware/forest/eris02_250702_D-2025-07-22-frame-at-0m24s.pdf}\hfill
  \includegraphics[width=0.245\linewidth,trim=265 100 290 1503,clip]{figures/hardware/forest/eris02_250702_D-2025-07-22-frame-at-0m26s.pdf}\hfill
  \includegraphics[width=0.245\linewidth,trim=265 100 290 1503,clip]{figures/hardware/forest/eris02_250702_D-2025-07-22-frame-at-0m27s.pdf}%
  \vspace{0.05cm}
  \includegraphics[width=0.9\linewidth]{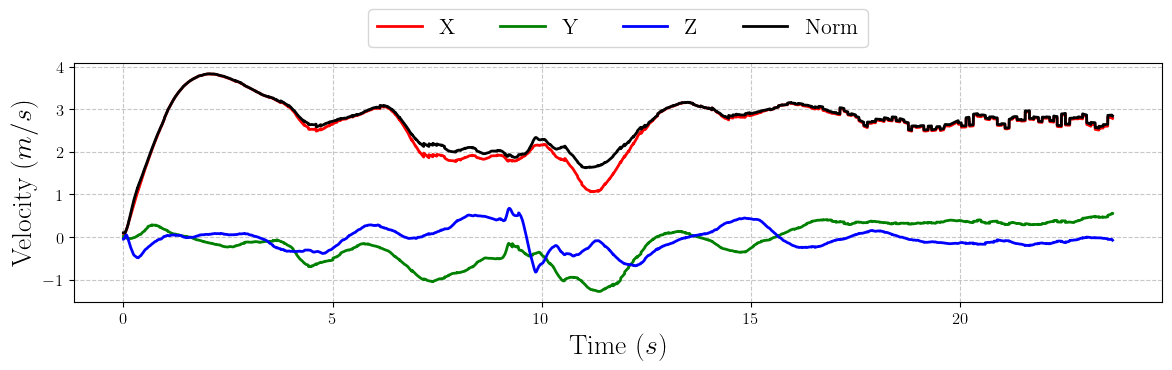}

  \caption{Outdoor obstacle avoidance test under tree canopy
  (\cref{tab:part2_hardware} Forest Flight 4).  The policy predicts up to
  \SI{30}{\degree} in yaw to navigate through dense underbrush with speeds up
  to \SI{3.8}{\meter\per\second}.  From top to bottom: VINS trajectory overlaid
  on terrain map, onboard RGB images, \red{inverted and max pooled depth images} policy depth input (inverted and max pooled), and
  velocity profile.
  }\label{fig:part2_forest_flight_d}%
\end{figure}

\begin{figure}
  \includegraphics[width=0.49\linewidth]{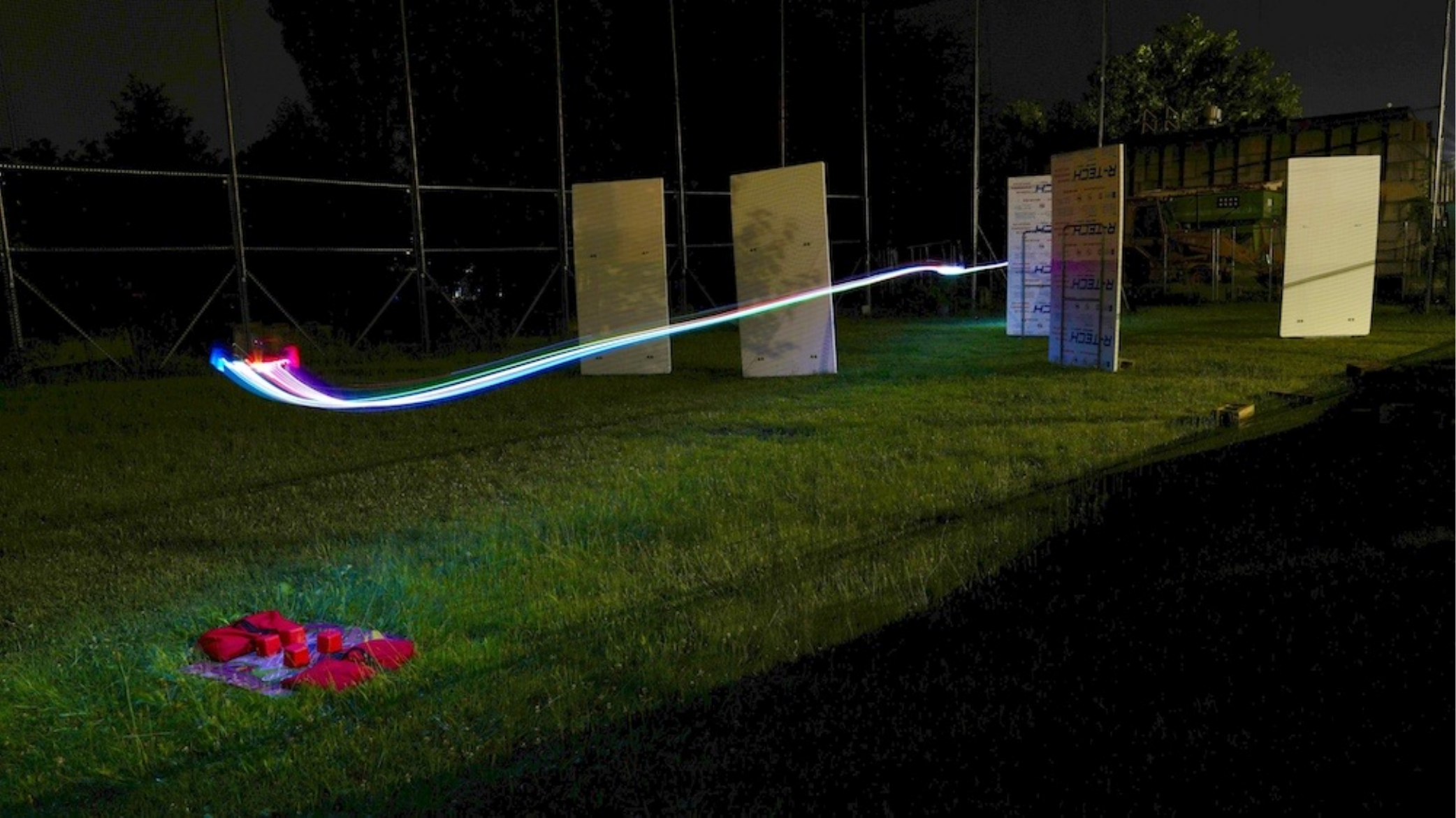}\hfill
  \includegraphics[width=0.49\linewidth]{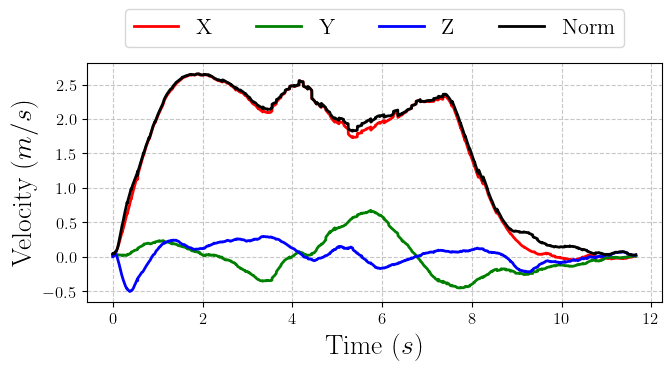}%
  \vspace{0.05cm}
  \includegraphics[width=0.49\linewidth]{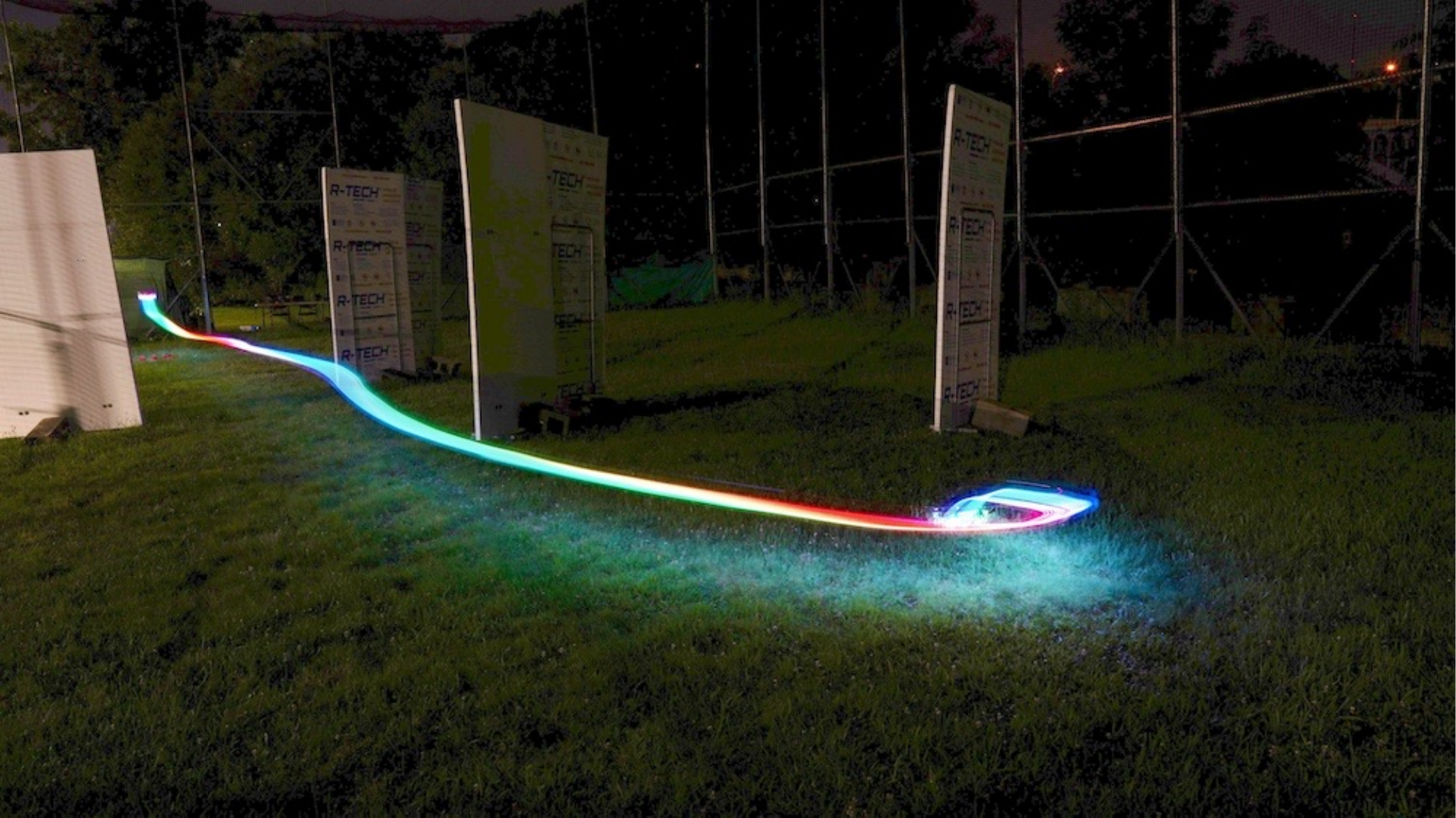}\hfill
  \includegraphics[width=0.49\linewidth]{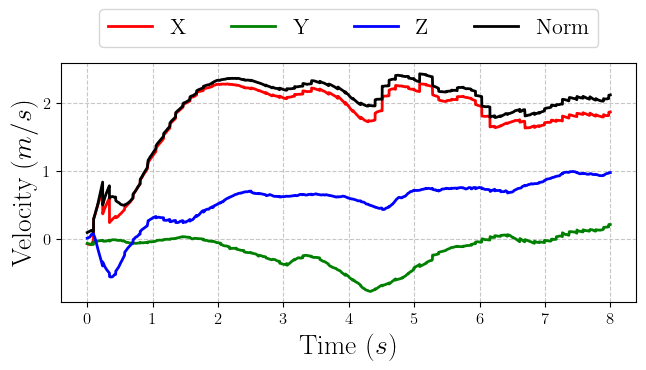}%
  \vspace{0.05cm}
  \includegraphics[width=0.49\linewidth]{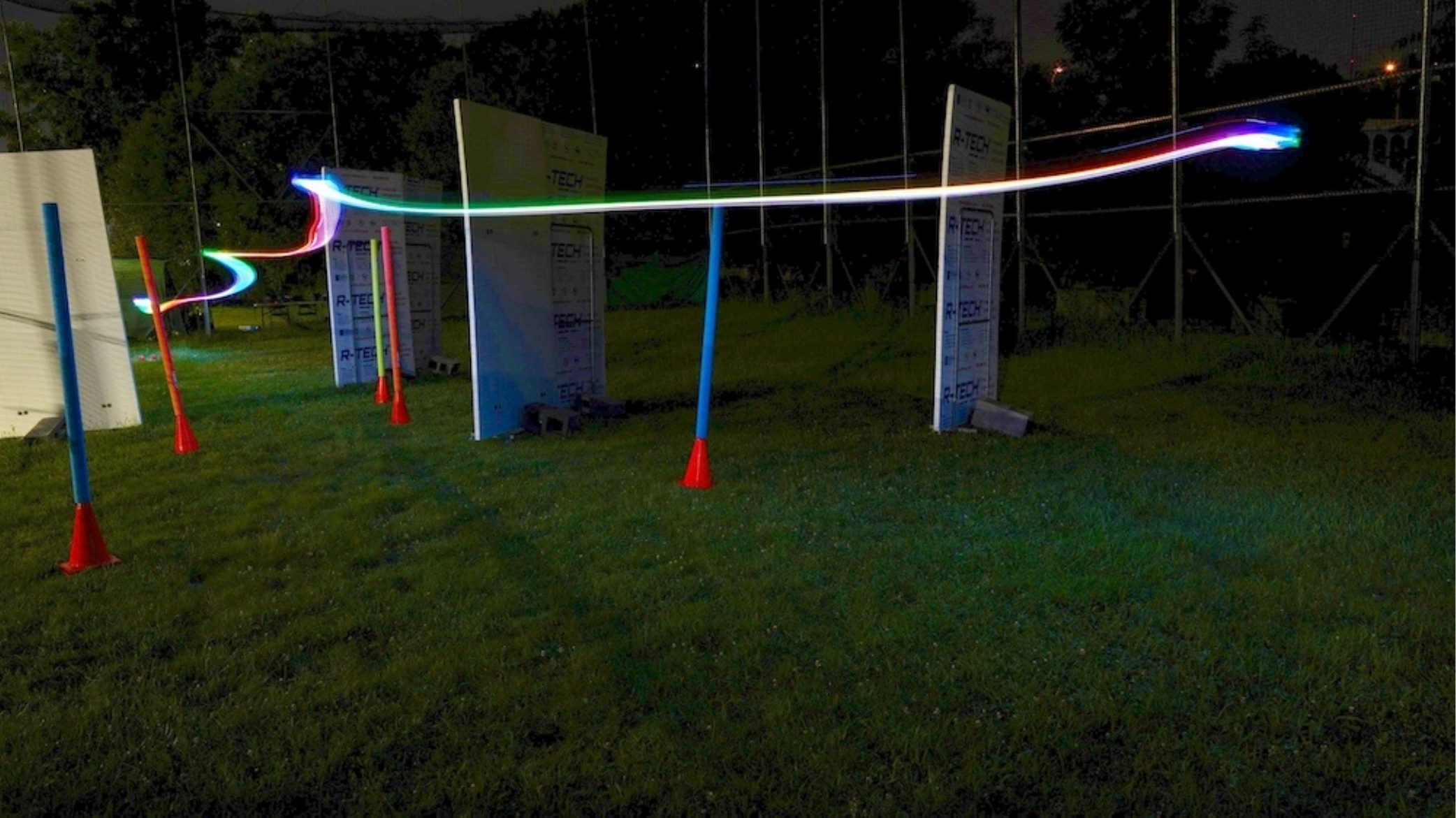}\hfill
  \includegraphics[width=0.49\linewidth]{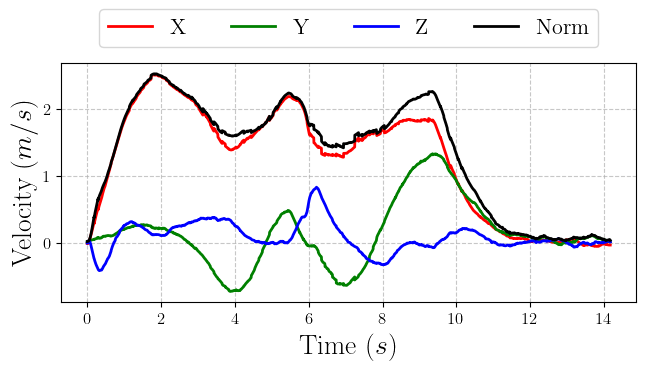}%
  \centering
  \caption{Outdoor obstacle avoidance tests at night using LED illumination and
  long-exposure photography.\label{fig:night_flight}}%
\end{figure}

We deploy the policies in hardware trials in an outdoor flight
arena (see~\cref{fig:night_flight}) and under a stand of trees with dense
foliage and underbrush (see~\cref{fig:part2_forest_flight_d}).
\Cref{tab:hardware_flights} provides a table of all flights conducted outdoors.
The top speed achieved by the system is
\SI{4}{\meter\per\second}. The total length over all trials is
\SI{589}{\meter}. No crashes occurred during testing.

\input{tables/hardware_trials.tex}

%% file: tables/hardware_trials.tex
\begin{table}[H]
\centering
\caption{Hardware Flight Trials\label{tab:hardware_flights}}
\label{tab:part2_hardware}
\resizebox{\columnwidth}{!}{
\begin{tabular}{llrrrrr}
\toprule
Env. & Flight & Flight Time & Path Length & $v_{max}$ & Max Speed & Avg Speed \\
& $\#$ & $s$ & $m$ & $m/s$ & $m/s$ & $m/s$ \\
\midrule
\multirow{14}{*}{\rotatebox{90}{Outdoor Flight Arena}}
    & 1 & 14.4 & 12.1 & 2.0 & 1.5 & 0.7 \\ 
    & 2 & 13.2 & 17.9 & 2.0 & 2.0 & 1.2 \\ 
    & 3 & 11.0 & 16.6 & 3.0 & 3.0 & 1.4 \\ 
    & 4 & 17.3 & 26.9 & 3.0 & 2.8 & 1.4 \\ 
    & 5 & 13.0 & 24.1 & 4.0 & 3.5 & 1.7 \\ 
    & 6 & 12.0 & 24.5 & 5.0 & 3.7 & 1.9 \\ 
    & 7 & 9.3 & 17.2 & 3.0 & 2.9 & 1.7 \\  
    & 8 & 11.7 & 22.3 & 3.0 & 2.9 & 1.8 \\ 
    & 9 & 11.8 & 24.9 & 3.0 & 4.0 & 2.0 \\ 
    & 10 & 11.1 & 22.5 & 3.0 & 2.8 & 1.8 \\ 
    & 11 & 11.7 & 19.6 & 3.0 & 2.7 & 1.5 \\ 
    & 12 & 11.3 & 20.4 & 3.0 & 2.7 & 1.6 \\ 
    & 13 & 10.0 & 19.8 & 3.0 & 2.9 & 1.9 \\ 
    & 14 & 9.5 & 17.9 & 3.0 & 2.4 & 1.9 \\  
    & 15 & 14.2 & 21.0 & 3.0 & 2.5 & 1.3 \\ 
\midrule
\multirow{5}{*}{\rotatebox{90}{Forest}}
    & 1 & 12.7 & 22.5 & 3.0 & 2.8 & 1.6 \\ 
    & 2 & 31.1 & 52.9 & 3.0 & 2.8 & 2.2 \\ 
    & 3 & 26.0 & 74.6 & 4.0 & 3.1 & 2.5 \\ 
    & 4 & 23.7 & 70.1 & 5.0 & 3.8 & 2.7 \\ 
    & 5 & 20.0 & 61.3 & 3.0 & 3.0 & 2.4 \\ 
\bottomrule
\end{tabular}
}
\end{table}

%% file: content/conclusion.tex
We developed a simulation and training framework for efficiently learning
vision-based navigation policies that map depth and state observations to thrust
and heading commands. Trained using simplified point-mass dynamics and
privileged information, the policy generalizes to full quadrotor dynamics and
reliably navigates through cluttered environments. Domain
randomization, especially over gravity, enables robust feedback control,
correcting for modeling errors such as a 15\% thrust mismatch. Our approach
achieves an 86\% success rate in simulation and completes 20 hardware flights
covering \SI{589}{\meter} without collisions.
While the policy generalizes from primitive training obstacles to
photorealistic simulation and real-world flights without fine-tuning, it
\red{While an improvement over the baselines, the proposed method} has difficulty
with backtracking in \red{dead ends} maze-like environments
(e.g., the \textit{Mine} scenario) and exhibits initial yaw oscillations. A
promising direction for future work is to explore more expressive learned-memory
architectures to enhance spatial reasoning and long-horizon planning without
relying on explicit maps.  Additionally, incorporating task-specific inputs and
new objective functions could extend the method's use to a wider range of
applications.

%% file: content/acknowledgements.tex
\ifthenelse{\equal{\blindmode}{true}}{
}{
\section*{ACKNOWLEDGMENTS}
The authors would like to thank Ankit Khandelwal for contributions to
the codebase and Edsel Burkholder for field testing support.
This material is based upon work supported in part by
the Army Research Laboratory and the Army Research Office under
contract/grant number W911NF-25-2-0153.
}